\documentclass[review]{elsarticle}

\usepackage{lineno,hyperref}
\modulolinenumbers[5]

\journal{}

\usepackage[utf8]{inputenc}
\usepackage{graphicx,amsfonts,amssymb,amsthm,stmaryrd,times}
\usepackage{amsmath}
\usepackage{graphicx,psfrag}
\usepackage[T1]{fontenc}
\usepackage[all]{xy}
\usepackage{multirow}
\usepackage{rotating}
\usepackage{xcolor}
\usepackage{dsfont}
\usepackage{wasysym}
\usepackage{url}
\usepackage{type1cm}
\usepackage{mathrsfs}
\usepackage{amsmath}
\usepackage{upgreek}
\usepackage{lettrine}
\theoremstyle{definition}
\newtheorem{thm}{Theorem}
\newtheorem*{thm*}{Theorem}

\newtheorem{deff}{Definition}
\newtheorem{exmp}{Example}
\newtheorem{rmk}{Remark}
\def\amax{\kern 0em\hbox{\rm \kern .25em\lower.1ex\hbox{\rlap{$\vee$}}\kern -.07em\lower.2ex\hbox{$\square$}\kern.25em}}
\def\amin{\kern 0em\hbox{\rm \kern .25em\lower.1ex\hbox{\rlap{$\wedge$}}\kern -.07em\lower.2ex\hbox{$\square$}\kern.25em}}

\def\boxmax{\kern 0em\hbox{\rm \kern .25em\lower.1ex\hbox{\rlap{$\vee$}}\kern -.07em\lower.2ex\hbox{$\square$}\kern.25em}}
\def\boxmin{\kern 0em\hbox{\rm \kern .25em\lower.1ex\hbox{\rlap{$\wedge$}}\kern -.07em\lower.2ex\hbox{$\square$}\kern.25em}}
\def\dualimp{\kern 0em\hbox{\rm \kern .25em\lower.1ex\hbox{\rlap{$\Rightarrow$}}\kern 0em\lower-1.2ex\hbox{$\overline{\hspace{2ex}}$}\kern.25em}}

\def\circmax{\kern 0em\hbox{\rm \kern .25em\lower.1ex\hbox{\rlap{$\vee$}}\kern -.18em\lower-.1ex\hbox{$\bigcirc$}\kern.25em}}
\def\circmin{\kern 0em\hbox{\rm \kern .25em\lower.1ex\hbox{\rlap{$\wedge$}}\kern -.18em\lower-.0ex\hbox{$\bigcirc$}\kern.25em}}

\newcommand{\vetx}{\mathbf{x}}
\newcommand{\vety}{\mathbf{y}}

\newcommand{\overrightarrowtx}{\mathbf{x}}

\newcommand{\sgn}{\text{sgn}}
\newcommand{\csgn}{\text{csgn}}
\newcommand{\qsgn}{\text{qsgn}}
\newcommand{\ii}{\mathbf{i}}
\newcommand{\jj}{\mathbf{j}}
\newcommand{\kk}{\mathbf{k}}

\newcommand{\re}[1]{\text{Re}\left\{#1\right\}}
\newcommand{\ve}[1]{\text{Ve}\left\{#1\right\}}

\newcommand{\bb}{\begin{equation}}
\newcommand{\ee}{\end{equation}}
\newcommand{\hyper}[2]{{#1}_0 + {#1}_1 \ii_1 + \ldots + {#1}_{#2} \ii_{#2}}

\newcommand{\obs}[1]{#1}

\begin{document}

\newpage
\thispagestyle{empty}
\begin{minipage}{0.8\textwidth}
\noindent {\Huge Copyright Notice} \vspace{2cm}

\noindent Personal use of this material is permitted.  Permission from Elsevier must be obtained for all other uses, in any current or future media, including reprinting/republishing this material for advertising or promotional purposes, creating new collective works, for resale or redistribution to servers or lists, or reuse of any copyrighted component of this work in other works.
\vspace{1cm}

\noindent Neural Networks. Volume 122, February 2020, Pages 54-67.  \url{https://doi.org/10.1016/j.neunet.2019.09.040}

\end{minipage}

\begin{frontmatter}

\title{A Broad Class of Discrete-Time Hypercomplex-Valued Hopfield Neural Networks}

\author{Fidelis Zanetti de Castro\corref{cor1}}
\ead{fidelis@ifes.edu.br}
\address{Federal Institute of Education, Science and Technology of Esp\'irito Santo at Serra, Rodovia ES-010, Km-6,5, Manguinhos, Serra-ES, CEP 29173-087, Brazil}

\author{Marcos Eduardo Valle\corref{cor2}}
\ead{valle@ime.unicamp.br}
\address{Department of Applied Mathematics, University of Campinas, Rua Sérgio Buarque de Holanda, 651, Campinas-SP, CEP 13083-859, Brazil}
\cortext[cor2]{Corresponding author}

\begin{abstract}
In this paper, we address the stability of a broad class of discrete-time hypercomplex-valued Hopfield-type neural networks. To ensure the neural networks belonging to this class always settle down at a stationary state, we introduce novel hypercomplex number systems referred to as real-part associative hypercomplex number systems. Real-part associative hypercomplex number systems generalize the well-known Cayley-Dickson algebras and real Clifford algebras and include the systems of real numbers, complex numbers, dual numbers, hyperbolic numbers, quaternions, tessarines, and octonions as particular instances. Apart from the novel hypercomplex number systems, we introduce a family of hypercomplex-valued activation functions called $\mathcal{B}$-projection functions. Broadly speaking, a $\mathcal{B}$-projection function projects the activation potential onto the set of all possible states of a hypercomplex-valued neuron. Using the theory presented in this paper, we confirm the stability analysis of several discrete-time hypercomplex-valued Hopfield-type neural networks from the literature. Moreover, we introduce and provide the stability analysis of a general class of Hopfield-type neural networks on Cayley-Dickson algebras.
\end{abstract}

\begin{keyword}
Hopfield neural network\sep hypercomplex-valued neural network\sep stability analysis\sep Clifford algebra\sep Cayley-Dickson algebra.
\end{keyword}

\end{frontmatter}


\section{Introduction} \label{sec:intro}
{Hopfield-type neural networks} (HNNs) are important recurrent neural networks that can be used to implement associative memories \citep{hopfield82}. Besides implementing associative memories, HNNs have been applied in control \citep{gan17,song17}, computer vision and image processing \citep{wang15,jli16}, classification \citep{pajares10,zhang17}, and optimization \citep{hopfield85,serpen08,cli16}. 

Although the HNN has been originally conceived for bipolar state neurons \citep{hopfield82}, it has been extended to hypercomplex-valued neurons using complex numbers \citep{jankowski96}, dual numbers \citep{kuroe11b}, hyperbolic numbers \citep{kobayashi13}, tessarines \citep{isokawa10}, quaternions \citep{isokawa08a}, octonions \citep{kuroe16}, and further hypercomplex number systems \citep{vallejo08,kuroe13,popa16c}. Extensions of the traditional HNN using hypercomplex numbers are referred to as {\it hypercomplex-valued Hopfield-type neural networks} (HHNNs). In this paper, we focus on {\it discrete-time} HHNNs.

Hypercomplex-valued neural networks can treat many kinds of information which may not be properly captured by real-valued neural networks, such as, phase, tensors, spinors, and multidimensional geometrical affine transformations \citep{fortuna96,nitta09,hirose12}. Moreover, in contrast to real-valued neural networks, hypercomplex-valued neural networks naturally treat multidimensional data as single entities. 
In other words, a hypercomplex-valued neural network can cope in an almost natural way with multidimensional data while constraints must be imposed on the topology of a real-valued neural network to take advantage of the correlation between the variables. In fact, a certain complex-valued neural network with a single hidden layer using frequency domain features outperformed the best real-valued neural network models in an image recognition task \citep{aizenberg18wcci}. Quaternion-valued neural networks outperformed real-valued networks for color and PolSAR image processing tasks \citep{minemoto16,shang14,kinugawa18}. Quaternion-valued neural networks also outperformed real-valued models for three- and four-dimensional time series prediction as well as target tracking \citep{greenblatt18,kumar18,xu16,talebi16}. 

Briefly, \obs{an HHNN} can be classified according to the activation function of its neurons. For example, a split HHNN is derived by applying a real-valued function to each component of the hypercomplex-valued activation potential of a neuron. Also, the target set of an activation function equals the set of all possible states of a hypercomplex-valued neuron. Accordingly, a multistate HHNN is obtained by considering a finite target set \citep{jankowski96,isokawa08a,isokawa13} while a continuous target set yields a continuous-valued HHNN \citep{aizenberg92,noest88b,noest88c,valle14bracis,valle17tnnls}. 

\subsection{A Short Literature Review on HHNNs}
Although research on discrete-time HHNN dates to the late 1980s \citep{aizenberg92,noest88b,noest88c}, in our opinion, the most relevant contribution is the complex-valued multistate model proposed by \cite{jankowski96}. The output of a multistate complex-valued neuron is obtained by applying the complex-valued signum function ($\csgn$) on the activation potential of a neuron. The $\csgn$ function, which has been introduced by \cite{aizenberg92}, yields the root of the complex unit which is obtained by quantizing the phase of its argument. \cite{jankowski96} have affirmed that their multistate complex-valued model with asynchronous update always \obs{settles} down at a stationary state under the usual conditions on the synaptic weights, that is,  Hermitian connections ($w_{ij} = \bar{w}_{ji}$), and {\it non-negative self-feedback} ($w_{ii} \geq 0$). However, \cite{zhou14} showed that the model of Jankowski et al. may fail to yield a convergent sequence if $w_{ii}=0$.  

In $2008$, Isokawa et al. extended the discrete-time multistate complex-valued neural network of Jankowski et al. to quaternions \citep{isokawa13,isokawa08b}. Briefly, a quaternionic version of the complex-valued signum function, denoted by $\qsgn$, is used in their model. The quaternionic signum function yields the unit quaternion obtained by quantizing the three phase-angles of its quaternionic argument. In $2016$, Minemoto et al. introduced a slight modification of the quaternionic multistate HNN of Isokawa et al. which is numerically stable \citep{minemoto16,valle16wcci}. Despite the successful application for the reconstruction of color images, the HHNN of Minemoto et al. does not always settle down at an equilibrium state \citep{valle17tnnls,valle16wcci}. Nevertheless, using a limit process, we obtain a discrete-time continuous-valued quaternionic HNN on unit quaternions which always comes to rest at an equilibrium state \citep{valle14bracis,valle17tnnls}. 

Apart from the extensions of the discrete-time HNN using complex numbers and quaternions, HHNNs on hyperbolic and dual numbers have been proposed and investigated by \cite{kobayashi13,kobayashi16c,kobayashi16d,kobayashihyperbolic17,kobayashi18b}. Also, \cite{castrocnmac17} introduced a discrete-time continuous-valued octonionic HHNN. HHNNs on tessarines, which are also referred to as commutative quaternions, have been investigated by \cite{isokawa10} and, more recently, by \cite{kobayashi18a}. 

\subsection{Contributions and Organization of the Paper}
It turns out that complex numbers, quaternions, and octonions are, apart from an isomorphism, all instances of Cayley-Dickson algebras. Complex numbers and quaternions are also isomorphic to instances of real Clifford algebras. In view of these remarks, this paper aims to provide a unifying framework for HHNNs defined on general hypercomplex number systems. 

Precisely, we introduce novel hypercomplex number systems which generalize the Cayley-Dickson and real Clifford algebras but enjoy the main properties for the stability analysis of HHNNs. The novel systems, referred to as \textit{real-part associative hypercomplex number systems}, provide one of the most general hypercomplex number systems for which a Hopfield-type neural network can be defined and properly analyzed.
We also introduce a broad family of hypercomplex-valued activation functions and provide an important theorem concerning the stability (in the sense of Lyapunov) for discrete-time hypercomplex-valued Hopfield-type neural networks. In fact, the theorem presented in this paper can be applied for the stability analysis of many discrete-time HHNNs from the literature, including complex-valued \citep{jankowski96,zhou14,kobayashicomplex17a}, hyperbolic-valued \citep{kobayashi13,kobayashi16c}, dual-numbered \citep{kobayashi18b}, tessarine-valued \citep{isokawa10,kobayashi18a}, quaternion-valued \citep{isokawa08a,valle17tnnls}, and octonion-valued Hopfield neural networks \citep{castrocnmac17}. 

In particular, this paper contributes with a slight modification of the complex-valued multistate Hopfield neural network of Jankowski et al. which always \obs{settle} down at an equilibrium state under the usual conditions on the synaptic weights, including $w_{ii} = 0$. In other words, using the theory developed for a broad class of discrete-time hypercomplex-valued Hopfield-type neural networks, in this paper we present a solution to an open problem in the literature. Moreover, we address the stability of complex-valued, hyperbolic-valued, and dual-numbered Hopfield-type neural networks with the split-sign function and we point out some interesting relations between them as well as their corresponding real-valued (bipolar) models. We also introduce a broad class of discrete-time Hopfield-type neural networks defined on Cayley-Dickson algebras which include the complex-valued, quaternion-valued, and octonion-valued models as particular instances. The stability analysis of the broad class of Cayley-Dickson Hopfield-type neural networks follows from the theory presented in this paper. 

This paper is organized as follows: Next section reviews the main concepts on hypercomplex numbers. The real-part associative hypercomplex number systems are also introduced in Section \ref{sec:numbers}. In Section \ref{sec:HHNN}, we introduce a broad class of activation functions and present a theorem concerning the stability of HHNNs. In Section \ref{sec:examples}, we apply the theory introduced previously for the stability analysis of several HHNNs from the literature. The modification of the complex-valued multistate Hopfield neural network of Jankowski et al. and the novel HHNNs on Cayley-Dickson algebras are presented in Section \ref{sec:examples}. Concluding remarks are given in Section \ref{sec:concluding}. 

\section{Hypercomplex numbers} \label{sec:numbers}
A number $p$ is said {\it hypercomplex} when it can be represented in the form
\bb \label{eq:hypercomplex} p= \hyper{p}{n}, \ee where $n$ is a non-negative integer, $p_0, p_1, \ldots, p_n$ are real numbers, and the symbols $\ii_1, \ii_2, \ldots, \ii_n$ (written using boldface in this paper) are called {\it hyperimaginary units} \citep{kantor89}. We denote by $\mathbb{H}$ the set (universe) of all hypercomplex numbers given by \eqref{eq:hypercomplex}. Examples of hypercomplex numbers include real numbers, complex numbers, hyperbolic numbers, dual numbers, quaternions, tessarines (also called commutative quaternions), and octonions, denoted in this paper respectively by $\mathbb{R}$, $\mathbb{C}$, $\mathbb{U}$, $\mathbb{D}$, $\mathbb{Q}$, $\mathbb{T}$, and $\mathbb{O}$. 

Note that a hypercomplex number $p=\hyper{p}{n}$ can be identified with the $(n+1)$-tuple $(p_0,p_1,\ldots,p_n)$ of real numbers. Thus, we say that the dimension of $\mathbb{H}$ is $n+1$ and write $\mathtt{dim}(\mathbb{H})=n+1$. For example, $\mathtt{dim}(\mathbb{R})=1$, $\mathtt{dim}(\mathbb{C})=\mathtt{dim}(\mathbb{U})=\mathtt{dim}(\mathbb{D})=2$, $\mathtt{dim}(\mathbb{Q})=\mathtt{dim}(\mathbb{T})=4$, and $\mathtt{dim}(\mathbb{O})=8.$ Furthermore, the set $\mathbb{H}$ of all hypercomplex numbers inherits the topology from $\mathbb{R}^{n+1}$. For instance, we say that $\mathcal{S} \subseteq \mathbb{H}$ is compact if and only if the set $\{(p_0,p_1,\ldots,p_n): \hyper{p}{n} \in \mathcal{S}\} \subseteq \mathbb{R}^{n+1}$ is compact. 

A hypercomplex number {\em system} is a set of hypercomplex numbers equipped with an {\em addition} and a {\em multiplication} (or \textit{product}). 
The addition of two hypercomplex numbers $p=\hyper{p}{n}$ and $q=\hyper{q}{n}$ is defined in a component-wise manner according to the expression 
\bb \label{eq:addition} p+q=(p_0+q_0)+(p_1+q_1)\ii_1+\ldots+(p_n+q_n)\ii_n. \ee 
The product between $p$ and $q$, denoted by the juxtaposition of $p$ and $q$, is defined as follows: First, we assign to each product of two hyperimaginary units $\ii_\mu$ and $\ii_\nu$, $\mu, \nu \in\{1,\ldots,n\}$, a new hypercomplex number. Mathematically, we define
\bb \ii_\mu\ii_\nu=a_{\mu\nu,0}+a_{\mu\nu,1}\ii_1+\ldots+a_{\mu\nu,n}\ii_n, \forall \mu,\nu \in \{1,\ldots, n\}. \label{eq:unitproduct}\ee 
We would like to point out that \eqref{eq:unitproduct} determines a {\it multiplication table}. Precisely, the product between $\ii_\mu$ and $\ii_\nu$ is the hypercomplex number situated in the intersection of the $\mu$th row and the $\nu$th column in this table. For instance, Table \ref{tab:multiplication} shows the multiplication table of the complex numbers, hyperbolic numbers, dual numbers, quaternions, and tessarines. For simplicity, we write $\ii_1=\ii,\ii_2=\jj$, and $\ii_3=\kk$.
\begin{table}
\begin{center}
\caption{Some multiplication tables} \label{tab:multiplication}
 \begin{tabular}{ccc}
  \footnotesize{a) Complex numbers} & \footnotesize{b) Hyperbolic numbers} & \footnotesize{c) Dual numbers} \\
 \begin{tabular}{c|c} 
  $\times$ & $\mathbf{i}$  \\ \hline 
  $\mathbf{i}$ & $-1$     
\end{tabular} &
\begin{tabular}{c|c} 
  $\times$ & $\mathbf{i}$  \\ \hline 
  $\mathbf{i}$ & $+1$     
\end{tabular} &
\begin{tabular}{c|c} 
  $\times$ & $\mathbf{i}$  \\ \hline 
  $\mathbf{i}$ & $0$    
\end{tabular} \\ \vspace{0.5em} 
\end{tabular}
%
\begin{tabular}{ccc}
\footnotesize{d) Quaternions} & & \footnotesize{e) Tessarines}\\
\begin{tabular}{c|c|c|c}
  $\times$ & $\mathbf{i}$ & $\mathbf{j}$ & $\mathbf{k}$ \\ \hline 
  $\mathbf{i}$ & $-1$  & $\mathbf{k}$ & $-\mathbf{j}$ \\ \hline
  $\mathbf{j}$ & $-\mathbf{k}$ & $-1$ & $\mathbf{i}$ \\ \hline
  $\mathbf{k}$ & $\mathbf{j}$ & $-\mathbf{i}$ & $-1$
\end{tabular} && 
\begin{tabular}{c|c|c|c}
  $\times$ & $\mathbf{i}$ & $\mathbf{j}$ & $\mathbf{k}$ \\ \hline 
  $\mathbf{i}$ & $-1$  & $\mathbf{k}$ & $-\mathbf{j}$ \\ \hline
  $\mathbf{j}$ & $\mathbf{k}$ & $+1$ & $\mathbf{i}$ \\ \hline
  $\mathbf{k}$ & $-\mathbf{j}$ & $\mathbf{i}$ & $-1$
\end{tabular}
\end{tabular}
\end{center}
\end{table}

The product $pq$ between the hypercomplex numbers $p$ and $q$ is determined using the {\it distributive law} and the {\it multiplication table} as follows. Each term $(p_\mu \ii_\mu)(q_\nu \ii_\nu)$ is rewritten as $p_\mu q_\nu(\ii_\mu \ii_\nu)$ and the product $\ii_\mu \ii_\nu$ is replaced in accordance with \eqref{eq:unitproduct}. Formally, the product is given by
\begin{align}
\label{eq:multiplication}
pq = \left( p_0q_0 + \sum_{\mu,\nu = 1}^n p_\mu q_{\nu} a_{\mu\nu,0}\right) &+ \left(p_0q_1+p_1q_0 + \sum_{\mu,\nu=1}^n p_\mu q_\nu a_{\mu\nu,1} \right) \ii_1 + \ldots  \nonumber \\
& + \left(p_0q_n+p_nq_0 + \sum_{\mu,\nu=1}^n p_\mu q_\nu a_{\mu\nu,n} \right) \ii_n. 
\end{align}
Note that we can identify a real number $\alpha \in \mathbb{R}$ with the hypercomplex number $\alpha+0\ii_1+\ldots+0\ii_n \in \mathbb{H}$. Hence, the scalar multiplication can be viewed as a particular case of the hypercomplex multiplication given by \eqref{eq:multiplication}.

The addition and the multiplication, given respectively by \eqref{eq:addition} and \eqref{eq:multiplication}, satisfy the following properties for all $\alpha,\beta \in \mathbb{R}$ and $p,q,r \in \mathbb{H}$:
\begin{enumerate}
\item $\alpha p= \alpha p_0+ (\alpha p_1){\ii_1}+\ldots+(\alpha p_n){\ii_n}=p\alpha$.
\item $(\alpha p)(\beta q)=(\alpha \beta)(pq)$.
\item $p(q+r)=pq+pr$ and $(p+q)r=pr+qr$.
\end{enumerate}

Many other properties of multiplication, such as commutativity and associativity, do not necessarily hold true. Indeed, the multiplication between tessarines is commutative and associative; the multiplication between quaternions is associative but it is not commutative; the multiplication between octonions is neither commutative nor associative.

A hypercomplex number $p$ can also be written as $p = p_0+\vec{p}$, where $p_0$ and $\vec{p}=p_1 \ii_1 + \ldots+p_n \ii_n$ are called, respectively, the real and the vector parts of $p$. We denote the real part of $p$ by $\re{p}:=p_0$ and its vector part by $\ve{p}:=\vec{p}$.

Borrowing the terminology of linear algebra, we speak of a linear operator $T:\mathbb{H} \to \mathbb{H}$ if $T(\alpha p + q) = \alpha T(p) + T(q)$, for all $p,q\in \mathbb{H}$ and $\alpha \in \mathbb{R}$. 
A linear operator that is an involution and also an antihomomorphism is called a {\em reverse-involution} \citep{ell07}, and it is formally defined as follows:

\begin{deff}[Reverse-involution]
An operator $\tau:\mathbb{H} \to \mathbb{H}$ is a reverse-involution if 
\begin{eqnarray} 
\label{eq:nu1}
&& \tau\big(\tau(p)\big) = p, \hfill\\
\label{eq:nu3}
&& \tau(pq)=\tau(q)\tau(p),\\
\label{eq:nu2}
&& \tau(\alpha p+q)=\alpha\tau(p)+\tau(q), 
\end{eqnarray}
for all $p,q \in \mathbb{H}$ and $\alpha \in \mathbb{R}$.  
\end{deff}
From \eqref{eq:nu1} and \eqref{eq:nu3}, we conclude that $1 \equiv 1+0\ii_1+ \ldots+0\ii_n$ is a fixed point of a reverse-involution $\tau$, that is, the equation $\tau(1)=1$ holds true. Furthermore, by writing $p = p_0 + \vec{p}$ and using the linearity property \eqref{eq:nu2}, we conclude that \[\tau(p) = \tau\big(p_0+\vec{p}\big) = \tau\big(p_0 \cdot 1 +\vec{p}\big)= p_0 \tau(1)+\tau(\vec{p}) = p_0 + \tau(\vec{p}).\] Therefore, we derive the important identity: 
\bb \label{eq:realtau} \re{\tau(p)}=\re{p}, \quad \forall p \in \mathbb{H}. \ee

The \textit{natural conjugation} is an example of a reverse-involution in some hypercomplex number systems such as the complex and quaternion number systems. Formally, the natural conjugate of a hypercomplex number $p$ is denoted by $\bar{p}$ and is defined by 
\bb \label{deff:conjug} \bar{p} = p_0-\vec{p}. \ee
Other examples of reverse-involutions include the \textit{quaternion anti-involutions} in quaternion algebra and the \textit{Clifford conjugation} in Clifford algebras \citep{ell07,delangue92,vaz16}. Also, if the multiplication is commutative, then the identity mapping $\tau(p)=p$, for all $p \in \mathbb{H}$, is referred to as the \textit{trivial reverse-involution}.

Finally, let us define the symmetric bilinear form $\mathcal{B}:\mathbb{H}\times \mathbb{H} \to \mathbb{R}$ by means of the following equation:
\bb \label{eq:inner-product} \mathcal{B}(p,q) = \re{\tau(p)q}, \quad \forall p,q \in \mathbb{H}.\ee
\begin{rmk}
The linearity in the first argument of $\mathcal{B}$ follows from \eqref{eq:nu2} while the following shows that it is also symmetric: 
\[\mathcal{B}(p,q) \mathop{=}^{\eqref{eq:inner-product}} 
\re{\tau(p)q} \mathop{=}^{\eqref{eq:realtau}} 
\re{\tau(\tau(p)q)} \mathop{=}^{\eqref{eq:nu3}} 
\re{\tau(q) \tau(\tau(p))} \mathop{=}^{\eqref{eq:nu1}} \re{\tau(q)p} \mathop{=}^{\eqref{eq:inner-product}} \mathcal{B}(q,p).\]
\end{rmk}
Intuitively, $\mathcal{B}$ measures a relationship between $p$ and $q$ by taking into account the algebraic properties of the multiplication and the reverse-involution $\tau$. For example, the symmetric bilinear form $\mathcal{B}$ coincides with the usual inner product on complex numbers, quaternions, and octonions with the natural conjugation.

\subsection{Real-Part Associative Hypercomplex Number Systems}
In general, the mathematical properties presented in the previous section hold true for an arbitrary hypercomplex number system $\mathbb{H}$. Let us now define a broad class of hypercomplex number systems with a reverse-involution that enjoy an extra property: associativity holds in the real-part of the product \citep{castrothesis18}. 

\begin{deff}[Real-Part Associative Hypercomplex Number Systems] \label{def1}
A hypercomplex number system equipped with a reverse-involution $\tau$ is called a \textit{real-part associative hypercomplex number system} (Re-AHN) if the following identity holds true for any three of its elements $p,q,r$:
\bb\re{(pq)r-p(qr)}=0. \label{eq:hopfieldtype}\ee
In particular, we speak of a {\it positive semi-definite (or non-negative definite)} real-part associative hypercomplex number system if the symmetric bilinear form $\mathcal{B}$ given by \eqref{eq:inner-product} satisfies
$\mathcal{B}(p,p) \geq 0$, $\forall p \in \mathbb{H}$.
\end{deff}

We would like to point out that the identity \eqref{eq:hopfieldtype} has been used implicitly for the stability analysis of many HHNN models \citep{jankowski96, isokawa13, valle14bracis}. Furthermore, this property is used explicitly when the product is not \obs{associative, for instance}, in the stability analysis of octonion-valued Hopfield-type neural networks \citep{kuroe16,castrocnmac17}. Finally, since the reverse-involution plays an important role for the stability analysis of HHNNs, it has been included in the definition of a real-part associative hypercomplex number system.

\begin{rmk} The element $(pq)r-p(qr)$ is called \textit{associator} of the real-part associative hypercomplex number systems relative to the product and measures its degree of non-associativity. Evidently, \eqref{eq:hopfieldtype} holds true if the multiplication is associative.
\end{rmk}

\begin{rmk} \label{rmk:inner-product}
In a real-part associative hypercomplex number system, the following identity holds true for any  $p,q,r \in \mathbb{H}$:
\bb \label{eq:inner-hermitian} \mathcal{B}(pq,r) = \mathcal{B}(q,\tau(p)r).\ee
\end{rmk}

We can prove by direct computation that the systems of real numbers, complex numbers, quaternions, and octonions are real-part associative hypercomplex number systems with the natural conjugation. More generally, real-part associative hypercomplex number systems include the tessarines, Cayley-Dickson algebras, and real Clifford algebras.

\begin{exmp} \label{ex:tessarines}
The set of \textit{tessarines}, also known as \textit{commutative quaternions} and denoted by $\mathbb{T}$, is composed by hypercomplex numbers of the form $p = p_0+p_1\ii+p_2\jj+p_3\kk$ whose multiplication is given by Table \ref{tab:multiplication}e). The reader interested on a historical account on the emergence of tessarines is invited to consult \citep{cerroni17}. The tessarines algebra is associative and, thus, the identity $\re{(pq)r} = \re{p(qr)}$ holds true for all $p,q,r \in \mathbb{T}$. Moreover, the reverse-involution given by
\bb \label{eq:tau_tessarines} \tau(p) = p_0 -p_1 \ii + p_2 \jj - p_3 \kk, \quad \forall p \in \mathbb{T},\ee
yields, by means of \eqref{eq:inner-product}, a symmetric bilinear form such that $\mathcal{B}(p,p)=p_0^2+p_1^2+p_2^2+p_3^2 \geq 0, \forall p \in \mathbb{T}$. Hence, $\mathbb{T}$ is a positive semi-definite real-part associative hypercomplex number system with the reverse-involution given by \eqref{eq:tau_tessarines}.
\end{exmp}

\begin{exmp}
Real Clifford algebras ($Cl(\mathbb{R})$), also referred to as geometric algebras, can be seen as hypercomplex number systems that incorporate the geometric notion of direction and orientation \citep{delangue92,vaz16,hestenes87}. Roughly speaking, a real Clifford algebra can be seen as a hypercomplex number system of dimension $n=2^k$ in which the product is associative and the square of an element is always a scalar, that is, 
\bb (pq)r = p(qr)  \quad \mbox{and} \quad p^2 \in \mathbb{R}, \quad \forall p,q,r \in Cl(\mathbb{R}). \ee 
The reader interested in the geometric interpretation as well as the formal definition of Clifford algebras is invited to consult \citep{delangue92,vaz16,hestenes87}. Also, a detailed survey on the applications of Clifford algebras with focus on neurocomputing and correlated areas can be found in \citep{hitzer13}.
In order to maintain geometric notions, the Clifford conjugation is a reverse-involution \citep{delangue92,vaz16}. Furthermore, since the identity \eqref{eq:hopfieldtype} holds true in an associative hypercomplex number system, we conclude that real Clifford algebras equipped with the Clifford conjugation are real-part associative hypercomplex number systems. Complex, hyperbolic, and dual numbers are all isomorphic to examples of 2-dimensional real Clifford algebras. Apart from an isomorphism, quaternion algebra is an example of a 4-dimensional real Clifford algebra.
\end{exmp}

\begin{exmp} \label{ex:Cayley1}
According to  \cite{biss07}, {\it Cayley-Dickson algebras} are finite-dimensional real algebras defined recursively as follows:
The first Cayley-Dickson algebra, denoted by $A_0$, is the real number system. Given a Cayley-Dickson algebra $A_k$, the next algebra $A_{k+1}$ comprises all pairs $(x,y) \in A_k \times A_k$ with the component-wise addition. The conjugation and product are defined as follows for any $(x_1,y_1), (x_2,y_2) \in A_{k+1}$:
\bb \label{eq:Cayley1} \overline{(x_1,y_1)} = (\bar{x}_1,-y_1),\ee
and
\bb \label{eq:Cayley2} (x_1,y_1)(x_2,y_2) = (x_1 x_2 - y_2 \bar{y}_1, \bar{x}_1y_2 + x_2 y_1).\ee
The real part is also defined recursively by setting $\re{x}=x$ for all $x \in A_0$ and
\bb \re{(x,y)} = \re{x}, \quad \forall (x,y) \in A_{k+1}, \ee
where the term on the right-hand side denotes the real part of $x \in A_k$. Also, the symmetric bilinear form $\mathcal{B}$ given by \eqref{eq:inner-product} satisfies $\mathcal{B}(x,x)=x^2$ for all $x \in A_0$ and   
\bb \label{eq:upsilon_cayley_recursive} \mathcal{B}\big((x_1,y_1),(x_2,y_2)\big) = \mathcal{B}(x_1,x_2)+\mathcal{B}(y_1,y_2)\ee for all $(x_1,y_1), (x_2,y_2) \in A_{k+1}.$

It is not hard to verify that $A_1$, $A_2$, and $A_3$ correspond respectively to the algebras of complex numbers, quaternions, and octonions. Furthermore, we can identify a Cayley-Dickson algebra $A_k$ with a positive semi-definite real-part associative hypercomplex number systems with dimension $2^k$. In fact, identity \eqref{eq:hopfieldtype} is a consequence of Lemma 2.8 from \cite{biss07}. By induction, it is not hard to show that the conjugation defined by \eqref{eq:Cayley1} corresponds to the natural conjugation given by \eqref{deff:conjug}. Also, the natural conjugation is a reverse-involution on $A_k$. Finally, from \eqref{eq:upsilon_cayley_recursive}, the symmetric bilinear form $\mathcal{B}:A_k\times A_k\to \mathbb{R}$ satisfies
\bb \label{eq:upsilon_cayley} \mathcal{B}(p,q) = \sum_{i=0}^{n} p_i q_i,\ee
for all $p=\hyper{p}{n}$ and $q=\hyper{q}{n}$ with $n=2^k-1$. Thus, $\mathcal{B}(p,p)=\sum_{i=0}^{2^k-1} p_i^2 \geq 0$, which implies that the Cayley-Dickson algebra $A_k$ is a positive semi-definite real-part associative hypercomplex number system.
\end{exmp}

\begin{figure}
\begin{center}
\includegraphics[width=0.7\columnwidth]{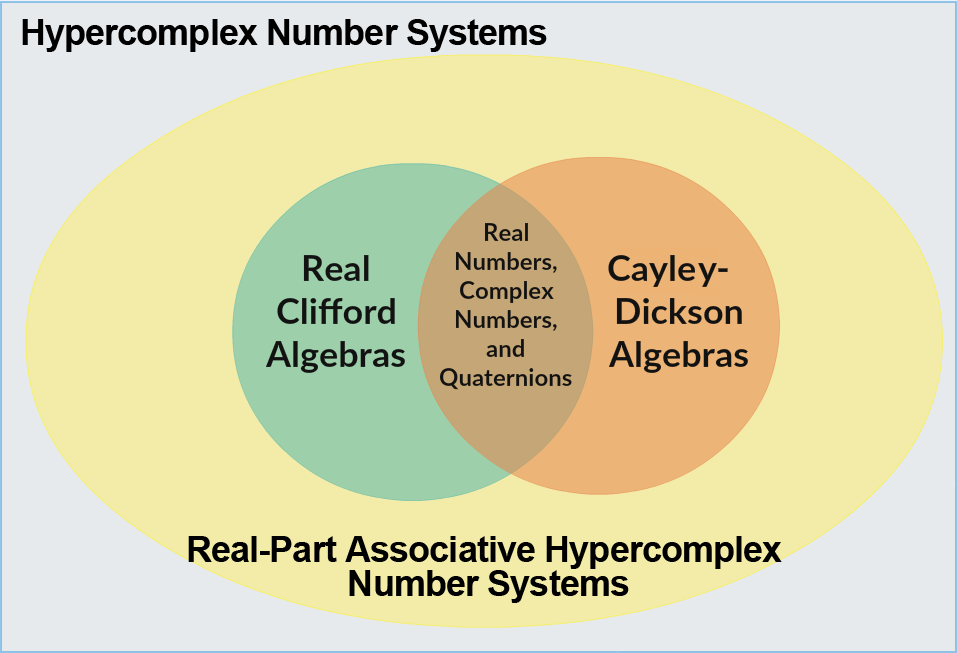}
\end{center}
\caption{Visual interpretation of the inclusion relationships between real-part associative hypercomplex number systems, Cayley-Dickson algebras eqquiped with the natural conjugation, and real Clifford algebras equipped with Clifford conjugation.
} \label{fig:diagram}
\end{figure}

Finally, the diagram shown in Figure \ref{fig:diagram} illustrates the inclusion relationships between some hypercomplex number systems.

\section{Hypercomplex Hopfield Neural Networks} \label{sec:HHNN}
Let $\mathbb{H}$ be a real-part associative hypercomplex number system and $\mathcal{S} \subset \mathbb{H}$ be the set of all possible states of a hypercomplex-valued neuron. Like the traditional discrete-time HNN, a discrete-time HHNN is a recurrent neural network with $N$ hypercomplex-valued neurons. Let $x_i(t) \in \mathcal{S}$ denote the hypercomplex-valued state of the $i$th neuron at time $t \geq 0$, for $i=1,\ldots,N$. Also, let $w_{ij} \in \mathbb{H}$ be the $j$th hypercomplex-valued synaptic weight of the $i$th neuron. Given an initial hypercomplex-valued state vector $\vetx(0) =[x_1(0),\ldots,x_N(0)]^T  \in \mathcal{S}^N$, the HHNN defines recursively the sequence $\{\vetx(t)\}_{t \geq 0}$ by means of the equation
\bb \label{eq:hopfield} x_i(t+\Delta t) = \begin{cases} f \big( v_i(t) \big), & v_i(t) \in \mathcal{D}, \\ x_i(t), &  \mbox{otherwise}, \end{cases} \ee
where $v_i(t)=\sum_{j=1}^{N}w_{ij} x_j(t)$ is the hypercomplex-valued activation potential of the $i$th neuron at time $t$, and $f$ is a hypercomplex-valued activation function with domain $\mathcal{D} \subset \mathbb{H}$ and codomain $\mathcal{S} \subset \mathbb{H}$. Note that the $i$th neuron remains in its state if the activation function is not defined at the hypercomplex-valued activation potential, that is, we have $x_i(t + \Delta t)=x_i(t)$ if $v_i(t) \not \in \mathcal{D}$. 

Broadly speaking, the activation function $f:\mathcal{D} \to \mathcal{S}$ in \eqref{eq:hopfield} should project the activation potential onto the set of all possible states of the hypercomplex-valued neuron. In other words, we expect $f(q)$ to be more related to $q \in \mathcal{D}$ than any other element $s \in \mathcal{S}$. Furthermore, this relationship between $f(q)$ and $q$ should take into account the algebraic structure of the real-part associative hypercomplex number system. Broadly speaking, $f$ should project $q$ onto $S$ with respect to the bilinear quadract form $\mathcal{B}$, which takes into account both the product and the reverse-involution. The following class of activation functions, which \obs{has} been introduced by \cite{castrothesis18}, formalizes these remarks. 

\begin{deff}[$\mathcal{B}$-projection function]
Consider a real-part associative hypercomplex number system $\mathbb{H}$ equipped with a reverse-involution $\tau$ and let $\mathcal{D} \subset \mathbb{H}$, $\mathcal{S} \subset \mathbb{H}$, and $\mathcal{B}:\mathbb{H} \times \mathbb{H} \to \mathbb{R}$ be the symmetric bilinear form defined by \eqref{eq:inner-product}. A hypercomplex-valued function $f: \mathcal{D} \to \mathcal{S}$ is called a \textit{$\mathcal{B}$-projection function} if
\bb \label{eq:class} \mathcal{B}(f(q),q) > \mathcal{B}(s,q), \quad  \forall q \in D, \forall s \in \mathcal{S} \setminus \left\lbrace f(q) \right\rbrace. \ee
\end{deff}

{
Examples of $\mathcal{B}$-projection functions include the complex-valued signum function for complex-valued Hopfield neural networks, the split-sign and the function that normalizes its arguments to length one on Cayley-Dickson algebras with the natural conjugation, and the split-sign functions for some real Clifford algebras with Clifford conjugation. Examples of $\mathcal{B}$-projection functions are addressed in details in Section 4. The multistate quaternion-valued signum function introduced by  \cite{isokawa08a} using quantizations of the phase-angles is not a $\mathcal{B}$-projection  function. 
At this point, however, we would like to call the reader’s attention to the following facts: First, note that $0 \equiv 0+0\ii_1+\ldots+0\ii_n$ cannot belong to the domain $D$ of a $\mathcal{B}$-projection function because $\mathcal{B}(p,0)=0$ for all $p \in \mathbb{H}$. Thus, the condition given by \eqref{eq:class} does not hold true if $0 \in D$. Second, but not less important, the inequality in \eqref{eq:class} depends not only on the expression of $f$ but also on the reverse-involution and the product of hypercomplex numbers (by means of the symmetric bilinear form).}

\subsection{Stability Analysis of HHNNs}
In many applications, including the implementation of associative memories and solving optimization problems, we are interested in the convergence of the sequence $\{\vetx(t)\}_{t \geq 0}$ generated by \obs{an HHNN}. In this subsection, we address this important issue. 
Precisely, we will study the dynamic of \obs{an HHNN} by means of the energy function $E$ defined by
\bb \label{eq:Energy} E(\vetx) = -\frac{1}{2} \displaystyle\sum_{i=1}^N\displaystyle\sum_{j=1}^N\re{\tau(x_i)(w_{ij}x_j)}, \ee where $N$ is the number of neurons of the neural network, $w_{ij}$ denotes the $j$th synaptic weight of the $i$th hypercomplex-valued neuron, and $\tau$ is the reverse-involution of a real-part associative hypercomplex number system $\mathbb{H}$. Alternatively, we can express the energy function $E$ as follows using the symmetric bilinear form $\mathcal{B}$ given by \eqref{eq:inner-product}:
\bb \label{eq:EnergyB} E(\vetx) = -\frac{1}{2} \sum_{i=1}^N \sum_{j=1}^N \mathcal{B}(x_i,w_{ij}x_j).\ee 

The convergence of the sequence $\{\vetx(t)\}_{t \geq 0}$ defined by \eqref{eq:hopfield} is ensured by showing that $E$ given by \eqref{eq:Energy} is real-valued, bounded and decreasing along any non-stationary trajectory, i.e, the inequality
\bb \Delta E = E(\vetx(t+\Delta t)) -  E(\vetx(t)) < 0,\ee
holds true whenever $\vetx(t+\Delta t) \neq \vetx(t)$.

Theorem \ref{thm:HHS}, whose proof can be found in the appendix, \obs{addresses} the convergence of the sequence produced by \eqref{eq:hopfield} with a $\mathcal{B}$-projection function $f$. 
\begin{thm} \label{thm:HHS}
Let $f:\mathcal{D} \to \mathcal{S}$ be a $\mathcal{B}$-projection function and $\vetx(0) \in \mathcal{S}^N$, where $\mathcal{S}$ is a compact subset of a real-part associative hypercomplex number system $\mathbb{H}$. The sequence produced by \eqref{eq:hopfield} is convergent, in an asynchronous update mode, if the synaptic weights satisfy $w_{ij}=\tau(w_{ji})$ and one of the two cases below holds true:
\begin{enumerate}[(a)]
\item $w_{ii}=0$ for any $i \in \{1,\ldots,N\}$. 
\item $w_{ii}$ {is a nonnegative real number for any $i \in \{1,\ldots,N\}$} and $\mathbb{H}$ is a positive semi-definite real-part associative hypercomplex number system.
\end{enumerate}
\end{thm}

\section{Examples} \label{sec:examples}
In this section we present examples illustrating that the theory presented in this paper generalizes several results from the literature concerning the stability analysis of HHNNs. We also use the theory presented above to introduce two families of Hopfield-type networks on Cayley-Dickson algebras (cf. Subsection \ref{sec:Cayley-Dickson}).

\subsection{Bipolar Hopfield Neural Network} \label{sec:pibolarHopfield}
Consider the system of real numbers $\mathbb{R}$ equipped with the {\em trivial reverse-involution}, i.e., the identity operator $\tau(x)=x$ for all $x \in \mathbb{R}$. Note that $\mathbb{R}$ with the identity operator is a positive semi-definite real-part associative hypercomplex number system because $\mathcal{B}(x,y) = xy$ and $\mathcal{B}(x,x) = x^2 \geq 0$. 

Apart from an isomorphism, the famous discrete-time real-valued recurrent neural network introduced by \cite{hopfield82} is given by \eqref{eq:hopfield} with the activation function $f \equiv \sgn:\mathcal{D} \to \mathcal{S}$, where the domain $\mathcal{D}=\mathbb{R}\setminus\{0\}$ is the set of all non-zero real numbers and $\mathcal{S} = \{-1,+1 \}$ is the set of all possible states of a real-valued bipolar neuron. The function $\sgn: \mathcal{D} \to \mathcal{S}$, which can be expressed as $\sgn(x)=x/|x|$, is a $\mathcal{B}$-projection function. In fact, for any $x \in \mathcal{D}$, we have $\mathcal{B}(\sgn(x),x) = |x|$ while  $s \in \mathcal{S} \setminus \{\sgn(x)\}$ implies $s = -\sgn(x)$ and, consequently, $\mathcal{B}(s,x) = - |x|$. Thus, $\mathcal{B}(\sgn(x),x) > \mathcal{B}(s,x)$ for all $x \in \mathcal{D}$ and $s \in \mathcal{S} \setminus \{\sgn(x)\}$. From Theorem \ref{thm:HHS}, the bipolar Hopfield neural network always settles down at an equilibrium if $w_{ij}=w_{ji}$ and $w_{ii}\geq 0$. 

This example highlights that the stability analysis of the traditional discrete-time Hopfield neural network is a particular case of the theory presented in this paper.

\subsection{Complex-Valued Multistate Hopfield Neural Networks} \label{sec:complexHopfield1}
As far as we know, although complex-valued Hopfield neural networks (CvHNNs) have been proposed in the late 1980s \citep{noest88b,noest88c}, the most relevant contribution is the multistate model proposed by \cite{jankowski96}. In fact, the multistate CvHNN of Jankowski et al. corroborated to the development of many other complex-valued Hopfield neural networks including the ones described on  \citep{kobayashicomplex17a,muezzinoglu03,lee06,tanaka09,kobayashicomplex17b,castrocnmac18,isokawa18}. 

Mistakenly, Jankowski et al. stated that their multistate CvHNN, operating asynchronously, always \obs{settles} down at an equilibrium state if the synaptic weight matrix satisfies the usual conditions: hermitian synaptic weights ($w_{ij}=\bar{w}_{ji}$) and non-negative self-connections ($w_{ii}\geq 0$). In 2014, Zhou and Zurada showed that the multistate CvHNN of Jankowski et al. may fail to yield a convergent sequence if $w_{ii}=0$ \citep{zhou14}. As pointed out by Zhou and Zurada, it turns out that the condition $w_{ii}=0$ is often used in applications of the multistate CvHNN. For instance, some design methods to implement an associative memory using multistate CvHNN, including the generalized projection rule \citep{lee06}, requires $w_{ii}=0$. Moreover, probably not aware of the subtle mistake by Jankowski et al., many multistate CvHNNs from the literature also fail to yield a convergent sequence of states under the usual conditions on the synaptic weights \citep{castrocnmac18}. Based on Theorem \ref{thm:HHS}, however, we provide below a solution to the stability analysis of multistate CvHNN under the usual conditions, including the case $w_{ii}=0, \forall i=1,\ldots,N$.

Consider the system of complex numbers $\mathbb{C}$ equipped with the natural conjugation given by \eqref{deff:conjug}. 
Note that the symmetric bilinear form $\mathcal{B}$ given by \eqref{eq:inner-product} satisfies
\bb \label{eq:complex_upsilon} \mathcal{B}(z_1,z_2) = |z_1| |z_2| \cos |\theta_1 - \theta_2|,\ee
where $z_1 = |z_1| e^{\ii \theta_1}$ and $z_2 = |z_2| e^{\ii \theta_2}$ are written in a polar representation such that $0 \leq |\theta_1-\theta_2| \leq \pi$.
In particular, $\mathcal{B}(z,z) = |z|^2 \geq 0$ and, thus, $\mathbb{C}$ with the natural conjugation is a positive semi-definite real-part associative hypercomplex number system.

Motivated by \cite{kobayashicomplex17a,kobayashicomplex17b}, let us define the complex-valued signum function as follows:
Given a positive integer number $K>1$, define $\Delta \theta=\pi/K$. The integer $K$ and the angle $\Delta \theta$ are referred respectively to as the \textit{resolution factor} and the \textit{phase-quanta}. The complex-valued signum function $\csgn: \mathcal{D} \to \mathcal{S}$ is defined by 
\bb \text{csgn}(z) = \begin{cases}
1, & 0 \leq \arg(z) < \Delta \theta, \\
e^{2\ii \Delta \theta}, & \Delta \theta < \arg(z) < 3 \Delta \theta, \\
\quad \vdots & \qquad \qquad \vdots \\
1, & (2K-1) \Delta \theta < \arg(z) < 2 \pi
             \end{cases} \label{eq:csgn}  
\ee
where
\bb \label{eq:SetD} \mathcal{D}=\{z \in \mathbb{C}\setminus \{0\}:\arg(z) \neq (2k-1) \Delta \theta, \forall k=1,\ldots,K \},\ee denotes its domain and  
\bb \label{eq:SetS} \mathcal{S}=\{1,e^{2\ii \Delta \theta}, e^{4 \ii \Delta \theta},\ldots, e^{ 2(K-1) \ii  \Delta \theta}\}\ee is the set of all possible states of a multistate complex-valued neuron. Note that $\csgn$ is not defined at a complex number $z$ such that $\arg(z)= (2k-1) \Delta \theta$, $k \in \{1,2,\ldots,K\}$. Therefore, from \eqref{eq:hopfield} with $f \equiv \csgn$, a neuron of a multivalued CvHNN is updated at time $t$ if and only if $v_i(t) \neq 0$ and $\arg(v_i(t)) \neq (2k-1) \Delta \theta$ for some $k \in \{1,2,\ldots,K\}$. This is the key issue to ensure that the sequences produced by a multistate CvMNN are all convergent. Precisely, in the following we show that $\csgn$ is a $\mathcal{B}$-projection function:
Given a complex number $z=|z|e^{\ii \theta} \in \mathcal{D}$, using the polar representation, let us write $\csgn(z) = e^{\alpha \ii}$ and $s = e^{\beta \ii}$ for any $s \in \mathcal{S} \setminus \{\csgn(z)\}$. From \eqref{eq:csgn}, we have 
\[ |\theta-\alpha|<\Delta \theta < |\theta-\beta|.\]
By applying the cosine function, multiplying by $|z|$, and using \eqref{eq:complex_upsilon}, we obtain
\bb \label{eq:csgn_hopfield} \mathcal{B}(\csgn(z),z)= |z| \cos|\theta-\alpha| >  |z| \cos|\theta-\beta| = \mathcal{B}(s,z).\ee
Concluding, from Theorem \ref{thm:HHS}, the complex-valued multistate Hopfield neural network with $f \equiv \csgn$ given by \eqref{eq:csgn} yields a convergent sequence if $w_{ij}=\bar{w}_{ji}$ and $w_{ii} \geq 0$ for all $i, j \in\{1,\ldots,N\}$. Moreover, in contrast to the many models in the literature, a neuron is updated if only if its activation potential $v_i(t) \in \mathcal{D}$, that is, $v_i(t) \neq 0$ and $\arg(v_i(t)) \neq (2k-1)\Delta \theta$ for some $k=1,\ldots,K$. At this point, note that the set $\mathbb{C}\setminus \mathcal{D}$ has Lebesgue measure zero. Thus, we almost always update a neuron using the $\csgn$ activation function. This remark justifies why the subtle mistake on the stability analysis of Jankowski et al. \obs{has} been unaware for almost 18 years. 

\subsection{Tessarine-Valued Hopfield Neural Networks} \label{sec:tessarines}
As far as we know, Isokawa et al. were the first to investigate Hopfield neural networks on tessarines, also known as commutative quaternions \citep{isokawa10}. Briefly, Isokawa and collaborators propose two tessarine-valued multistate discrete-time Hopfield neural networks based on polar representations of tessarines. Although we believe that both tessarine-valued Hopfield neural networks of Isokawa et al. can be analyzed using the theory presented previously in this paper, to simplify our discussion, in the following we only address a slight modification of the tessarine-valued multistate Hopfield neural network introduced recently by \cite{kobayashi18a} and which is very similar to the second model proposed by \cite{isokawa10}. 

Consider the hypercomplex number system $\mathbb{T}$ of tessarines equipped with the reverse-involution $\tau$ given by \eqref{eq:tau_tessarines}, i.e., $\tau(p) = p_0 - p_1\ii+p_2 \jj - p_3 \kk$ for all $p = p_0+p_1\ii+p_2\jj+p_3 \kk \in \mathbb{T}$. In this case, $\mathbb{T}$ is a positive semi-definite real-part associative hypercomplex number system and the symmetric bilinear form given by \eqref{eq:inner-product} satisfies $\mathcal{B}_\mathbb{T}(p,q) = p_0 q_0 + p_1q_1 +p_2q_2 + p_3q_3$ for all $p,q \in \mathbb{T}$. 
Since $\kk = \ii\jj$, a tessarine can be written as the direct sum of two complex numbers $u_p = p_0 + p_1\ii$ and $v_p = p_2+p_3\ii$ as follows:
\bb \label{eq:bicomplex} p=p_0+p_1\ii+p_2\jj+p_3\kk = (p_0+p_1\ii)+(p_2+p_3\ii)\jj = u_p + v_p\jj.\ee
Moreover, because $\mathcal{B}_\mathbb{C}(z_1,z_2)=a_1a_2 + b_1b_2$ for any complex-numbers $z_1= a_1 +b_1\ii$ and $z_2=a_2+b_2\ii$, the symmetric bilinear form satisfies the following identity for any two tessarines $p=u_p+v_p\jj$ and $q=u_q+v_q\jj$:
\bb \label{eq:upsilon_tessarines2complex} \mathcal{B}_\mathbb{T}(p,q) = \mathcal{B}_\mathbb{C}(u_p,u_q) + \mathcal{B}_\mathbb{C}(v_p,v_q).\ee

In analogy to the complex-valued multistate Hopfield neural network, given a positive integer number $K>1$ called \textit{resolution factor}, define the \textit{phase-quanta} by means of the equation $\Delta \theta=\pi/K$. Using \eqref{eq:bicomplex}, Kobayashi defines the set of all possible states of a tessarine-valued multistate neuron as follows:
\bb \label{eq:tessarineS} \mathcal{S}_\mathbb{T}=\{ p = u_p + v_p\jj: u_p,v_p \in \mathcal{S}\},\ee 
where $\mathcal{S}=\{1,e^{2\ii \Delta \theta}, e^{4 \ii \Delta \theta},\ldots, e^{2(K-1) \ii  \Delta \theta}\} \subset \mathbb{C}$. Also, in accordance with \cite{kobayashi18a}, let us define the tessarine-valued activation function $\text{tsgn}: \mathcal{D}_\mathbb{T} \to \mathcal{S}_\mathbb{T}$ by means of the equation 
\bb \text{tsgn}(p) = \text{csgn}(u_p)+\text{csgn}(v_p)\jj,  \quad \forall p=u_p+v_p\jj \in \mathcal{D}_\mathbb{T}, \ee
where $\text{csgn}$ denotes the complex-valued activation function given by \eqref{eq:csgn} and 
\bb \mathcal{D}_\mathbb{T} = \{p = u_p + v_p\jj: u_p,v_p \in \mathcal{D}\}.\ee 
For any $p=u_p+v_p\jj \in \mathcal{D}_\mathbb{T}$ and $s = u_s+v_s \jj \in \mathcal{S}_\mathbb{T} \setminus \{\text{tsgn}(p)\}$, from \eqref{eq:csgn_hopfield} and \eqref{eq:upsilon_tessarines2complex}, we obtain 
\begin{align*} 
\mathcal{B}_\mathbb{T}(s,p) &
= \mathcal{B}_\mathbb{C}(u_s,u_p)+\mathcal{B}_\mathbb{C}(v_s,v_p) \\ 
&< \mathcal{B}_\mathbb{C}(\csgn(u_p),u_p)+\mathcal{B}_\mathbb{C}(\csgn(v_p),v_p) \\ &= \mathcal{B}_\mathbb{T}(\text{tsgn}(p),p). 
\end{align*}
Therefore, the tessarines $\mathbb{T}$ with the reverse-involution $\tau$ defined by \eqref{eq:tau_tessarines} is a positive semi-definite real-part associative hypercomplex number system and $\text{tsgn}:\mathcal{D}_\mathbb{T} \to \mathcal{S}_\mathbb{T}$ is a $\mathcal{B}$-projection function. From Theorem \ref{thm:HHS}, we conclude that the tessarine-valued multistate Hopfield neural network described by \eqref{eq:hopfield} yields a convergent sequence in an asynchronous update mode if the synaptic weights satisfy $w_{ij} = \tau(w_{ji})$ and $w_{ii}\geq 0$ for all $i,j=1,\ldots,N$. 

Concluding, the theory presented in this paper provides a unified mathematical explanation for the stability analysis of many hypercomplex-valued neural networks such as the one detailed by \cite{isokawa10} and \cite{kobayashi18a}. 

\subsection{Some Clifford-Valued Hopfield Neural Networks} \label{sec:clifford}
As far as we know, Vallejo and Bayro-Corrochano provided the first account on discrete-time Hopfield neural networks on Clifford algebras \citep{vallejo08}. Apart from the complex-valued HHNN, which is a real Clifford algebra of dimension equal to $2$,  Hopfield neural networks on hyperbolic and dual domains have been investigated by \cite{kobayashi13,kobayashi16c,kobayashi16d,kobayashihyperbolic17,kobayashi18a}. Moreover, continuous-time Hopfield neural networks on Clifford algebras have been extensively investigated by \cite{kuroe11a,kuroe13,kuroe11b}. In this subsection, we apply the theory presented in this paper to confirm the stability analysis of Hopfield neural networks on some real Clifford algebras. Precisely, we shall focus on Hopfield neural networks based on the so-called split-sign function and defined on real Clifford algebras of dimension 2. 

A real Clifford algebra of dimension 2 corresponds to an associative hypercomplex number system with elements of the form $p=p_0+p_1\ii$ where the square of the hypercomplex unit is a real number, that is, $\ii^2 \in \mathbb{R}$. As a consequence, a real Clifford algebra of dimension 2 is isomorphic to either the systems of complex numbers ($\mathbb{C}$), hyperbolic numbers ($\mathbb{U}$), or dual numbers ($\mathbb{D}$). We recall that the multiplication table of these three hypercomplex number systems are given by Table \ref{tab:multiplication}a), b), and c). Moreover, it is not hard to show that the systems of complex, hyperbolic, and dual numbers are all commutative algebras. 

The Clifford conjugation corresponds to the natural conjugation on $\mathbb{C}$, $\mathbb{U}$, and $\mathbb{D}$. Apart from the natural conjugation, the identity mapping is also a reverse-involution on the complex, hyperbolic, and dual number systems. In view of this remark, it is convenient to introduce the notation
\[ \tau_\lambda(p) = p_0+\lambda p_1\ii, \quad \forall p=p_0+p_1\ii,\]
where $\lambda \in \{-1,+1\}$. Note that $\tau_\lambda$ is the trivial reverse-involution if $\lambda=+1$ while, when $\lambda=-1$, we obtain the natural conjugation. In both cases, $\tau_\lambda$ is a reverse-involution on $\mathbb{C}$, $\mathbb{U}$, and $\mathbb{D}$. Furthermore, complex, hyperbolic, and dual number systems with $\tau_\lambda$ are all real-part associative hypercomplex number systems. The symmetric bilinear form given by \eqref{eq:inner-product} satisfies the following equation for all $p=p_0+p_1\ii$ and $q=q_0+q_1\ii$:
\bb \label{eq:upsilon_clifford2} \mathcal{B}(p,q) = p_0 q_0 + \lambda p_1q_1 \ii^2, \ee
where $\lambda \in \{-1,+1\}$ and $\ii^2 \in \{-1,0,+1\}$. 
In particular, we have $\mathcal{B}(p,p) \geq 0$ for all $p = p_0+p_1\ii$ if and only if $\lambda \ii^2 \geq 0$. Thus, the system of complex numbers with the natural conjugation is a positive semi-definite real-part associative hypercomplex number system. Similarly, the system of hyperbolic numbers yields a positive semi-definite real-part associative hypercomplex number system with the identity mapping. Finally, the system of dual numbers is a positive semi-definite real-part associative hypercomplex number system with both the identity and the natural conjugation.

Now, consider the split-sign function $\sgn:\mathcal{D} \to \mathcal{S}$ defined in a component-wise manner by means of the equation
\bb \label{eq:sgn2} \sgn(p) = \sgn(p_0)+\sgn(p_1)\ii, \quad \forall p = p_0+p_1\ii,\ee
where 
\bb \nonumber \mathcal{D} = \{p = p_0 + p_1\ii: p_0p_1 \neq 0\} \ee
and
\bb \label{eq:DSsgn2} \mathcal{S} = \{1+\ii,1-\ii,-1-\ii,-1+\ii\}.\ee
It can be easily verified that the inequality 
\[ \mathcal{B}(\sgn(q),q) = |q_0| + \lambda |q_1| \ii^2 >  q_0 s_0 + \lambda q_1 s_1 \ii^2 = \mathcal{B}(s,q),\]
holds true for all $q=q_0+q_1\ii \in \mathcal{D}$ and $s=s_0+s_1\ii \in \mathcal{S}\setminus\{\sgn(q)\}$ if and only if $\lambda \ii^2\geq 0$. From Theorem \ref{thm:HHS}, we conclude that a sequence produced by a Hopfield neural network with the split-sign function is convergent if $\lambda \ii^2 \geq 0$, $w_{ij}=\tau_\lambda(w_{ji})$, and $w_{ii}\geq0$ for all $i,j=1,\dots, N$. In other words, a Hopfield neural network given by \eqref{eq:hopfield} with the split-sign function always yields a convergent sequence in the asynchronous update mode when:

\noindent \textbf{-- Complex numbers:} The synaptic weights satisfy $w_{ij} = \bar{w}_{ji}$ and $w_{ii} \geq 0$ for all $i,j=1,\ldots,N$. 

\noindent \textbf{-- Hyperbolic numbers:} The synaptic weights satisfy, for all $i=1,\ldots,N$, the conditions $w_{ij}=w_{ji}$ and $w_{ii} \geq 0$. In a similar manner, the theory presented in this paper can be applied for the stability analysis of the two hyperbolic-valued neural networks proposed by  \cite{kobayashi16c,kobayashi18a}. 

\noindent \textbf{-- Dual numbers:} The synaptic weights satisfy $w_{ii} \geq 0$ and either $w_{ij} = \bar{w}_{ji}$ or $w_{ij} = w_{ji}$ holds true for all $i,j=1,\ldots,N$. Accordingly, \cite{kobayashi18b} asserts that the identity $\re{w_{ij}} = \re{w_{ji}}$ ensures the stability of this recurrent neural network with $w_{ii}=0$ for all $i=1,\ldots,N$.

Another interesting result that emerges from the theory presented in this paper is obtained by considering the function $\overline{\sgn}:\mathcal{D} \to \mathcal{S}$ defined by
\bb \label{eq:sgn3} \overline{\sgn}(p) = \sgn(p_0)-\sgn(p_1)\ii, \quad \forall p = p_0+p_1\ii,\ee
where $\mathcal{D}$ and $\mathcal{S}$ are given by \eqref{eq:DSsgn2}. Note that $\overline{\sgn}(p) = \sgn(\bar{p}) = \overline{\sgn(p)}$. Therefore, the dynamic of the Hopfield neural networks on 2-dimensional real Clifford algebras with $\overline{\sgn}$ corresponds to the recurrent neural networks obtained by considering $\sgn$ but either conjugating the activation potential or the output of the neuron.
In contrast to Hopfield neural networks on 2-dimensional real Clifford algebras with the $\sgn$ function, the inequality
\bb \label{eq:upsilon_sgn3} \mathcal{B}(\overline{\sgn}(q),q)>\mathcal{B}(s,q), \quad \forall q \in \mathcal{D}, s \in \mathcal{S}\setminus\{\overline{\sgn}(q)\}, \ee
holds true if and only if $\lambda \ii^2\leq 0$. For example, \eqref{eq:upsilon_sgn3} holds if $\ii^2=-1$ and $\lambda=1$. In other words, $\overline{\sgn}$ is a $\mathcal{B}$-projection function on the complex numbers equipped with the trivial reverse-involution. Although $\mathbb{C}$ with the identity mapping is a real-part associative hypercomplex number system, it is not positive semi-definite. Thus, from Theorem \ref{thm:HHS}, we conclude that a complex-valued Hopfield neural network with the activation function \eqref{eq:sgn2} and asynchronous update always comes to rest at an equilibrium if the synaptic weights satisfy $w_{ij}=w_{ji}$ and $w_{ii}=0$. Accordingly, we have just derived the stability conditions for the symmetric complex-valued Hopfield neural network proposed by \cite{kobayashicomplex17a} and further discussed using the split-sign activation function by \cite{kobayashi17b}. Similarly, we conclude that \obs{an HHNN} given by \eqref{eq:hopfield} with the $\overline{\sgn}$ activation function always \obs{settle} down at a stationary state in the asynchronous update mode if:

\noindent \textbf{-- Complex numbers:}  The synaptic weights satisfy $w_{ii}=0$ and $w_{ij} = {w}_{ji}$ for all $i,j=1,\ldots,N$. 

\noindent \textbf{-- Hyperbolic numbers:} The synaptic weights satisfy $w_{ii}= 0$ and $w_{ij}=\bar{w}_{ji}$ for all $i,j=1,\ldots,N$. 

\noindent \textbf{-- Dual numbers:} The synaptic weights satisfy $w_{ii} \geq 0$ and either $w_{ij} = \bar{w}_{ji}$ or $w_{ij} = w_{ji}$ holds true for all $i,j=1,\ldots,N$.

Let us conclude this subsection with two simple illustrative examples.
\begin{exmp} \label{ex:Clifford}
Consider the four-state Hopfield neural network with $N=2$ neurons on the three real Clifford algebras of dimension 2 and synaptic weights given by
 \[ w_{11} = w_{22}= 0 \quad \mbox{and} \quad w_{12}=w_{21}=1+3\ii.\]
Note that this Hopfield neural network has symmetric weights and no self-feedback. Thus, the complex-valued neural network with $\overline{\sgn}$ activation function, the hyperbolic-valued neural network with $\sgn$ activation function, and both dual-numbered neural networks always come to rest at an equilibrium. In contrast, the complex-valued neural network with ${\sgn}$ and the hyperbolic-valued neural network with $\overline{\sgn}$ may fail to settle down at a stationary state. Indeed, Figure \ref{fig:graph} shows the directed graphs obtained from the six Hopfield neural networks on the real Clifford algebras of dimension 2. In these directed graphs, a node corresponds to a state of the neural network while an edge from node $i$ to node $j$ means that we can obtain the $j$th state from the $i$th state by updating a single neuron. {For illustractive purposes, we depicted with red edges all the possible trajectories obtained from the initial state $\vetx(0) = [-1-\ii,1+\ii]$, which corresponds to the state number four in the directed graphs.} Note from Figure \ref{fig:graph} that there is a one-to-one correspondence between the a) complex-valued HNN with $\overline{\sgn}$ and b) the hyperbolic-valued HNN with the $\sgn$. Although these two HHNN models settle down at different equilibrium states, they exhibit very similar dynamics. Similarly, there is also a one-to-one correspondence between e) the complex-valued HNN with $\sgn$ and f) the hyperbolic-valued HNN with the $\overline{\sgn}$. In particular, these two models exhibit limit cycles which \obs{prevent} the HNNs to settle down at a stationary state. Finally, observe that the dynamic of the dual-numbered HNN with $\sgn$ activation function is similar to the dynamic of the dual-numbered HNN with $\overline{\sgn}$ -- both models have the same number of equilibrium points. 
Concluding, apart from an application of the theory presented in this paper for stability analysis of several models from the literature, this example reveals an interesting relationship between the Hopfield neural networks on real Clifford algebras of dimension 2 which requires further study.
 
 \begin{figure}
 \begin{center}
 \begin{tabular}{c}
  \footnotesize{a) Complex-valued Hopfield neural network with $\overline{\sgn}$.} \\
  \includegraphics[width=0.9\linewidth]{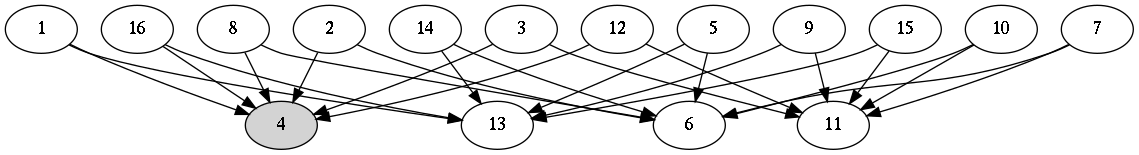} \\
  \footnotesize{b) Hyperbolic-valued Hopfield neural network with $\sgn$.} \\
  \includegraphics[width=0.9\linewidth]{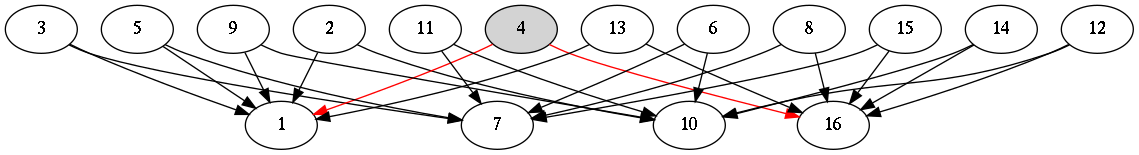} \\
  \footnotesize{c) Dual-numbered Hopfield neural network with $\sgn$.} \\
  \includegraphics[width=0.7\linewidth]{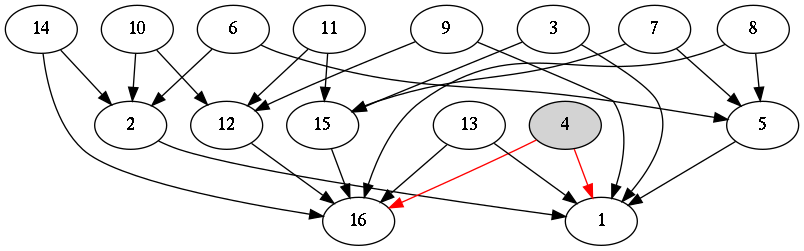} \\ 
  \footnotesize{d) Dual-numbered Hopfield neural network with $\overline{\sgn}$.} \\
  \includegraphics[width=0.7\linewidth]{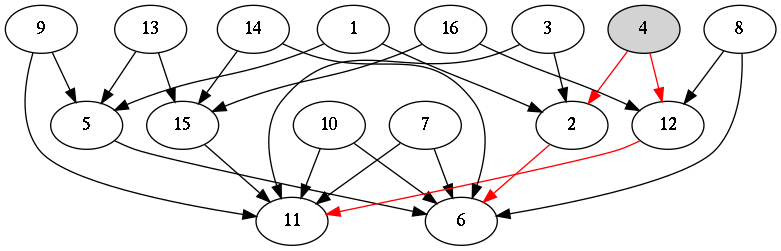} \\
  \begin{tabular}{cc}
    \footnotesize{e) Complex-valued Hopfield neural network with $\sgn$.} &
    \footnotesize{f) Hyperbolic-valued Hopfield neural network with $\overline{\sgn}$.} \\
    \includegraphics[width=0.45\linewidth]{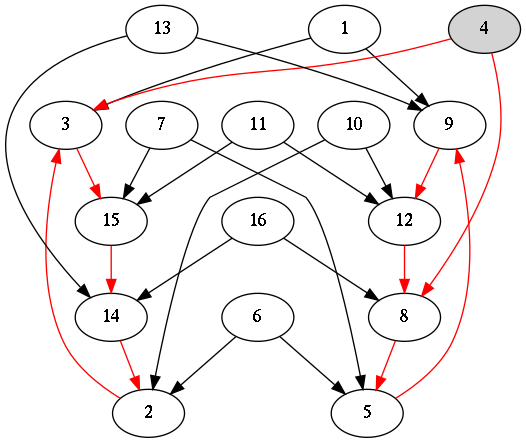} &
    \includegraphics[width=0.45\linewidth]{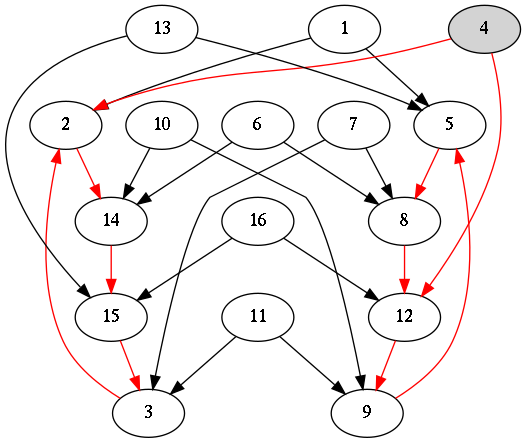}
  \end{tabular} \\
 \end{tabular}
 \caption{Dynamic of the six hypercomplex-valued Hopfield neural networks considered in Example \ref{ex:Clifford}. Possible limit cycles are marked in red.} \label{fig:graph}
 \end{center}
 \end{figure}
\end{exmp}

\begin{exmp} \label{ex:Clifford_real}
{The previous example addressed the dynamic of HHNNs on real Clifford algebras of dimension 2. Let us now compare these HHNNs with the split-sign activation function with their corresponding real-valued versions. Precisely, a complex number $a_1+b_1\ii$, a hyperbolic number $a_2+b_2\ii$, and a dual-number $a_3+b_3\ii$ can be identified, respectively, with the following $2\times 2$-matrices: 
\[\begin{bmatrix} a_1 & -b_1\\b_1 & a_1 \end{bmatrix}, \quad \begin{bmatrix} a_2 & b_2\\b_2 & a_2 \end{bmatrix} \quad \mbox{and} \quad \begin{bmatrix} a_3 & 0\\b_3 & a_3 \end{bmatrix}.\] 
As a consequence, the complex-valued, hyperbolic-valued, and dual-numbered Hopfield neural networks with the split-sign activation function of Example \ref{ex:Clifford} can be \obs{identified} with real-valued bipolar Hopfield neural networks with the split-sign activation function and four neurons $(N=4$) whose synaptic weight matrices are given respectively by 
 \[ W^c = {\small \left[ \begin{array}{cc|cc}
            0 & 0  & 1 &  -3 \\
            0 &  0  & 3 &  1 \\
            \hline 
            1  & -3  & 0 &  0 \\
            3  & 1 &  0 &  0
           \end{array}
 \right]}, 
 W^h = {\small \left[ \begin{array}{cc|cc}
            0 & 0  & 1 &  3 \\
            0 &  0  & 3 &  1 \\
            \hline 
            1  & 3  & 0 &  0 \\
            3  & 1 &  0 &  0
           \end{array}
 \right]}, \mbox{ and } 
 W^d = {\small \left[ \begin{array}{cc|cc}
            0 & 0  & 1 &  0 \\
            0 &  0  & 3 &  1 \\
            \hline 
            1  & 0  & 0 &  0 \\
            3  & 1 &  0 &  0
           \end{array}
 \right]}. \]
Note that only $W^h$ is symmetric with non-negative diagonal. Hence, the real-valued neural network derived from the symmetric hyperbolic-valued neural network always \obs{settles} down at an equilibrium point. From the classical result for bipolar Hopfield neural network \citep{hopfield82}, however, nothing can be said about the other two real-valued neural networks. The directed graphs depicted on Figure \ref{fig:graphreal} illustrate the dynamic of the three real-valued Hopfield neural networks. In analogy to Figure \ref{fig:graph}, all possible trajectories obtained by starting the neural network at $\vetx(0)=[-1-\ii,1+\ii]$, which corresponds to the state number four, have been depicted using red edges. Note that the real-valued neural network obtained from the complex-valued Hopfield neural network also \obs{exhibits} limit cycles. Precisely, in contrast to the complex-valued neural network, its corresponding real-valued Hopfield neural network may exhibit a chaotic behavior because any state can be reached from any initial state. Although the hypercomplex-valued Hopfield neural network and its corresponding real-valued neural network have the same stationary states, they exhibit different dynamics. Indeed, the hyperbolic-valued Hopfield neural network settles at an equilibrium with at most one single neuron update. The real-valued Hopfield neural network, however, may require two updates to reach a stationary state. Furthermore, starting at state number four, the corresponding real-valued neural network can reach any of the four possible stationary states (follow the red edges in the graph in Figure \ref{fig:graphreal}b)). In contrast, starting at the same initial state, the hyperbolic-valued HNN can settle only on two of the four stationary states (see the red edges in the graph in \obs{Figure} \ref{fig:graph}b)). In particular, the hyperbolic neural network has 50\% of chance to settle at the stationary state number 1 while the real-valued as only 25\% of chance to settle at the same stationary state. Putting this remark in a practical context, if neural networks are designed to implement associative memories \citep{hopfield82,hassoun97}, the real-valued Hopfield neural network is more likely to settle at a spurious memory than the hyperbolic-valued neural network. Finally, the dual-numbered Hopfield neural network and its corresponding real-valued neural network also have the same stationary states. The real-valued Hopfiled neural network, however, usually require more updates to settle at an equilibrium than the dual-numbered model. Concluding, this example suggests that hypercomplex-valued Hopfield neural networks may be simpler (e.g. in terms of the complexity of the resulting directed graphs) than their corresponding real-valued models. In particular, in this example the hypercomplex-valued models usually required fewer updates to settle down at an equilibrium state than their corresponding real-valued neural networks.}

 \begin{figure}[t]
 \begin{center}
 \begin{tabular}{cc}
   \parbox[t]{0.45\columnwidth}{ 
  {\footnotesize{a) Real-valued Hopfield neural network obtained from the complex-valued Hopfield neural network.}} \\
  \includegraphics[width=0.9\linewidth]{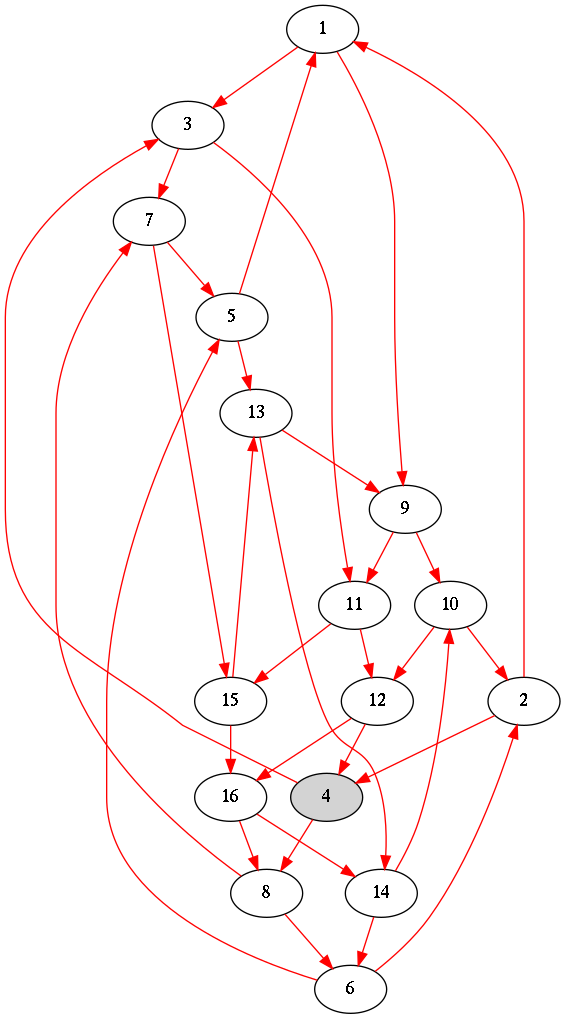} } &
  \parbox[t]{0.54\columnwidth}{
  {\footnotesize{b) Real-valued Hopfield neural network derived from the hyperbolic-valued Hopfield neural network.}} \\
  \includegraphics[width=1\linewidth]{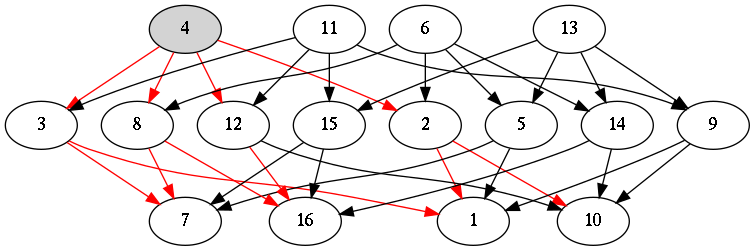} \\ \vspace{1em} \\
  {\footnotesize{c) Real-valued Hopfield neural network obtained from the dual-numbered Hopfield neural network.}} \\
  \includegraphics[width=0.9\linewidth]{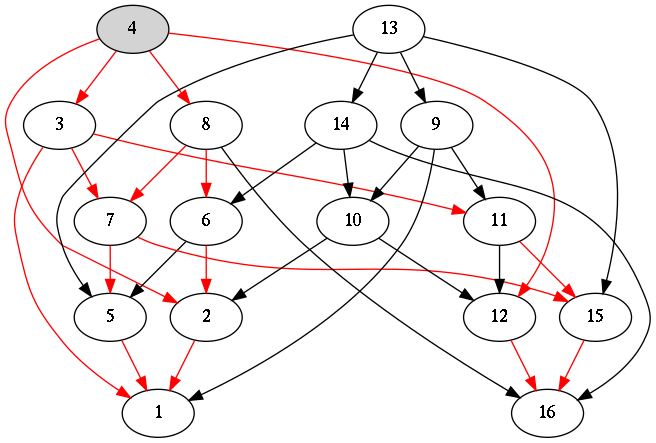} }
 \end{tabular}
   \caption{Dynamic of the real-valued Hopfield neural networks obtained from a) the complex-valued Hopfield neural network, b) the hyperbolic-valued Hopfield neural network, and c) the dual-numbered Hopfield neural network of Example \ref{ex:Clifford}.} \label{fig:graphreal}
  \end{center}
 \end{figure}
 
 \end{exmp}

Apart from the models in the literature, the following subsection reveals that the theory presented in this paper can be used to study the dynamic of novel HHNN models.

\subsection{HHNNs on Cayley-Dickson Algebras} \label{sec:Cayley-Dickson}

In analogy to the multistate CvHNN, a continuous-valued CvHNN is obtained by considering in \eqref{eq:hopfield} the activation function defined by $\sigma(z)=z/|z|$ for all $z \neq 0$. The complex-valued discrete-time Hopfield neural network obtained by considering $\sigma$ as the activation function has been investigated by \cite{noest88c}. More recently, the continuous-valued CvHNN \obs{has} been extended to quaternions independently by \cite{valle14bracis} and \cite{kobayashiX}. We also used this kind of activation function to introduce a discrete-time continuous-valued octonionic Hopfield neural network \citep{castrocnmac17}. At this point, we would like to recall that Kuroe and Iima have investigated the stability of continuous-time octonionic Hopfield neural networks \citep{kuroe16}. Their study motivated us to investigate the discrete-time models and inspired us to introduce the theory presented in this paper. Interestingly, the complex, quaternionic, and octonionic continuous-valued Hopfield neural networks, operating asynchronously, yield a convergent sequence of states if the synaptic weights satisfy the usual conditions: $w_{ij}=\bar{w}_{ji}$ and $w_{ii}\geq0$ for all $i,j=1,\ldots,N$. Since the systems of complex numbers, quaternions, and octonions are all instances of Cayley-Dickson algebras, let us apply the theory developed in this paper for the analysis of the stability of HHNN models defined on an arbitrary Cayley-Dickson algebra $A_k$.

As pointed out previously, a Cayley-Dickson algebra $A_k$ whose conjugation and product are defined recursively by \eqref{eq:Cayley1} and \eqref{eq:Cayley2} with $A_0 =\mathbb{R}$ is a positive semi-definite real-part associative hypercomplex number system. Although the multiplication may \obs{fail} to have some desirable algebraic properties such as commutativity and associativity, the Cayley-Dickson algebra $A_k$ enjoys some properties from Euclidean geometry. Precisely, from \eqref{eq:upsilon_cayley}, the symmetric bilinear form $\mathcal{B}:A_k \times A_k \to \mathbb{R}$ 
corresponds to the usual inner product between $p=\hyper{p}{n} \equiv (p_0,p_1,\ldots,p_n)$ and 
$q=\hyper{q}{n} \equiv (q_0,q_1,\ldots,q_{n})$, where $n=2^k-1$.
Moreover, the absolute value of a hypercomplex number $p\equiv (p_0,p_1,\ldots,p_{n}) \in A_k$ can be defined by 
\bb |p| = \sqrt{\mathcal{B}(p,p)} = \sqrt{\sum_{i=0}^{2^k-1} p_i^2},\ee
which corresponds to the Euclidean norm.
From the Cauchy-Schwarz inequality, we have $\mathcal{B}(p,q) \leq |p||q|$, with equality if and only if $q = \alpha p$ for some real $\alpha>0$. 

In analogy to the complex-valued, quanternionic, and octonionic continuous-valued HHNNs, let $\sigma: \mathcal{D} \to \mathbb{S}$ be the activation function given by 
\bb \label{eq:sigma} \sigma(p)=\frac{p}{|p|}, \quad \forall p \in \mathcal{D}=A_k \setminus \{0\},\ee
where $\mathbb{S}=\{p \in A_k:|p|=1\}$. The following shows that $\sigma$ is a $\mathcal{B}$-projection function. Consider $p \in \mathcal{D}$ and $s \in \mathcal{S} \setminus \{\sigma(p)\}$. Since $s \neq \alpha p$, for any $\alpha>0$, we obtain
\[ \mathcal{B}(s,p) <|s||p|=|p| = \mathcal{B}(\sigma(p),p).\] 
The inequality is a consequence of the Cauchy-Schwarz inequality and the fact that $p$ and $s$ are not parallel vectors.
Thus, $\mathcal{B}(\sigma(p),p)>\mathcal{B}(s,p)$ for all $p \in \mathcal{D}$ and $s \in \mathbb{S} \setminus \left\lbrace \sigma(p) \right\rbrace$. Since $A_k$ is a positive semi-definite real-part associative hypercomplex number system, from Theorem \ref{thm:HHS}, the sequences generated by \eqref{eq:hopfield} with $f \equiv \sigma$ are all convergent if the synaptic weights satisfy the usual conditions $w_{ij}=\bar{w}_{ji}$ and $w_{ii}\geq 0$. 

{
In analogy to the continuous-valued Hopfield neural networks, it is \obs{rather} straightforward to extend the real-valued, complex-valued, and quaternion-valued Hopfield neural networks with split-sign activation function to Cayley-Dickson algebras. Precisely, consider a Cayley-Dickson algebra $A_k$ and let the set of states of a neuron be \[S = \{p = \hyper{p}{n} \in A_k: p_\mu \in \{-1,+1\}, \forall \mu=0,1,\ldots,n\},\] 
where $n=2^k-1$. Also, consider the split-sign activation function $\sgn:\mathcal{D} \to \mathcal{S}$ defined in a component-wise manner as follows for all $q \in \mathcal{D}$, where $\mathcal{D} = \{q \in A_k: q_0 q_1 \cdots q_n \neq 0\}$: 
\bb \label{eq:CayleySplit} \sgn(q) = \sgn(q_0) + \sgn(q_1)\ii_1 + \ldots + \sgn(q_n) \ii_n. \ee
Geometrically, the split-sign function projects its argument onto a vertex of a hypercube of a $2^k$-dimensional space.
Since the symmetric bilinear form $\mathcal{B}:A_k\times A_k \to \mathbb{R}$ corresponds to the usual inner product, the following holds true for any $q \in \mathcal{D}$ and $s \in \mathcal{S}\setminus\{\sgn(q)\}$: 
\bb \label{eq:CayleyFunction} \mathcal{B}(s,q) = \sum_{\mu=0}^n s_\mu q_\mu < \sum_{\mu=0}^n |q_\mu| = \sum_{\mu = 0}^n \sgn(q_\mu)q_\mu = \mathcal{B}(\sgn(q),q). \ee
The inequality \eqref{eq:CayleySplit} follows because there exists an index $\nu \in \{0,1,\ldots,n\}$ such that  $s_\nu = -1$ if $q_\nu>0$ and $s_\nu=+1$ if $q_\nu<0$. Equivalently, $s_\nu q_\nu = -|q_\nu|$ for some  $\nu$. From \eqref{eq:CayleyFunction}, we conclude that the split-sign is also a $\mathcal{B}$-projection function on a Cayley-Dickson algebra $A_k$. Furthermore, from Theorem \ref{thm:HHS}, the sequences generated by \eqref{eq:hopfield} with $f \equiv \sgn$ are all convergent if the synaptic weights satisfy the usual conditions $w_{ij}=\bar{w}_{ji}$ and $w_{ii}\geq 0$. 
}

\begin{exmp} 
{Let us illustrate the dynamic of an octonion-valued (Cayley-Dickson algebra $A_3$) Hopfield neural network with the split-sign activation function and compare it with its corresponding real-valued bipolar Hopfield neural network. Precisely, we synthesized an octonion-valued Hopfield neural network with $N=100$ neurons and weights defined as follows for $i=1,\ldots,N$ and $j=i+1,\ldots,N$: 
\[ w_{ii} = 0, \quad w_{ij} = \mathtt{randn}+ \mathtt{randn} \ii_1 + \ldots + \mathtt{randn} \ii_7, \quad \mbox{ and } \quad w_{ji} = \bar{w}_{ij}.\]
Here, $\mathtt{randn}$ yields a real number sampled from a normal probability distribution with mean 0 and standard deviation 1. The corresponding real-valued neural network is obtained by considering a bipolar Hopfield neural network with $8N$ neurons and real-valued synaptic weight matrix $M$ such that $M \phi(\vetx) = \phi(W \vetx)$, where $\phi:\mathbb{O}^N \to \mathbb{R}^{8N}$ is the bijection defined as follows for any $\vetx  = [(\hyper{x_1}{7}),\ldots,(\hyper{x_N}{7})] \in \mathbb{O}^N$: 
\[\phi(\vety) = [{x_{1}}_0,\ldots,{x_{1}}_7,{x_{2}}_0,\ldots,{x_{2}}_7,\ldots,{x_{N-1}}_0,\ldots,{x_{N-1}}_7,{x_{N}}_0,\ldots,{x_{N}}_7]^T. \]
In words, the multiplication of the real-valued synaptic weight matrix $M$ by the real-valued vector $\phi(\vetx)$ obtained by concatenating all the entries of the octonion-valued vector $\vetx$ corresponds to the real-valued vector obtained by concatenating the entries of the product between the octonion-valued matrix $W$ and $\vetx$ \obs{\citep{tian00}}. Furthermore, the hypercomplex-valued neural network \obs{has} been initialized at an octonion-valued vector $\vetx(0) \in \mathcal{S}^N$ whose components $x_i(0)={x_i}_0(0) + {x_i}_1(0)\ii_1 + \ldots + {x_i}_7(0)\ii_7$ are uniformly distributed in $\mathcal{S}$, that is, $\mbox{Pr}[{x_i}_\mu(0) = +1] = 0.5$ for all $i=1,\ldots,N$ and $\mu = 0,\ldots,7$. The real-valued Hopfield neural network \obs{has} been initialized with the $8N$-dimensional bipolar vector $\phi(\vetx(0))$ obtained by concatenating all the entries of the octonion-valued input. The evolution of the energy given by \eqref{eq:Energy} for both octonion-valued and real-valued Hopfield neural network from one outcome of this experiment is shown in Fig. \ref{fig:Energy}. Both neural networks evolved until they reach a stationary state. \obs{Note that the octonion-valued as well as the real-valued model settled at a stationary state whose energy value is close to $-16000$. Nevertheless, }
the octonion-valued Hopfield neural network reached a steady state with much less updates than its corresponding real-valued neural network. Intuitively, the fast convergence of the octonion-valued neural network follows because one hypercomplex-valued neuron can be viewed as a group of 8 real-valued neurons. Furthermore, we may conjecture that hypercomplex-valued Hopfield neural networks overcome real-valued models when a certain entity can be described by a hypercomplex number or a group of real-valued neurons. In the future, we intend to develop efficient learning strategies for hypercomplex-valued Hopfield neural networks and further compare their performance with real-valued models.
}

\begin{figure}[t]
 \begin{center}
  \includegraphics[width = 0.9\linewidth]{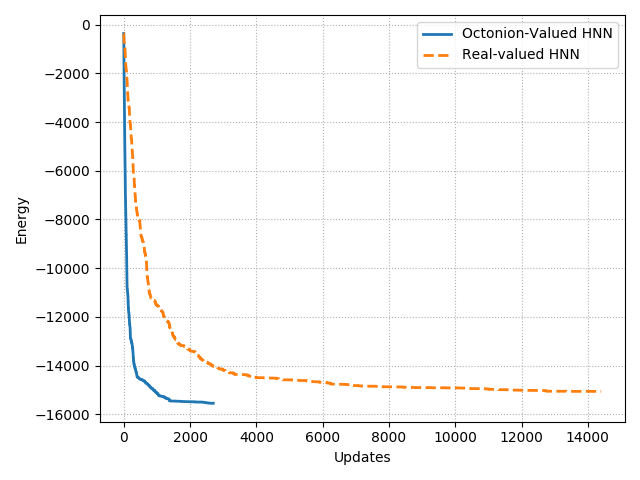}
 \end{center} 
 \caption{Evolution of the energy of an octonion-valued Hopfield neural network with split-sign activation function and its corresponding real-valued bipolar neural network by the number of neuron updates.} \label{fig:Energy}
\end{figure}
\end{exmp}

Concluding, the stability analysis of the complex-valued \citep{noest88c}, quaternionic \citep{valle14bracis, valle17tnnls, kobayashiX}, and octonionic-valued \citep{castrocnmac17} Hopfield neural networks can be derived as a particular case of the theory presented in this paper. More generally, this theory can be applied in a straightforward manner to the broad class of continuous-valued and split-sign Hopfield neural networks defined on a Cayley-Dickson algebra $A_k$.

\section{Concluding Remarks} \label{sec:concluding}
In this paper, we addressed the stability of a broad class of discrete-time hypercomplex-valued Hopfield-type neural networks (HHNNs). To this end, we introduced new hypercomplex number systems called real-part associative hypercomplex number systems. Real-part associative hypercomplex number systems provide an appropriate mathematical background for the development of HHNNs and include, as particular instances, Cayley-Dickson algebras and real Clifford algebras. 

Apart from the new hypercomplex number systems, in this paper we also introduced a broad family of hypercomplex-valued functions, referred to as $\mathcal{B}$-projection functions. The stability of \obs{an HHNN} with a $\mathcal{B}$-projection function is ensured by means of Theorem \ref{thm:HHS} under mild conditions on the synaptic weights. It should be emphasized that the results presented in this paper extend several results published in the literature on the stability analysis of discrete-time Hopfield-type neural networks since the early 1980s. Moreover, it can be applied for the development of many new HHNN models. Indeed, we used the theory presented in this paper to introduce a broad class of HHNNs on Cayley-Dickson algebras. 

{Finally, we would like to recall that the real-valued Hopfield neural network \obs{has} been applied in control \citep{gan17,song17}, computer vision and image processing \citep{wang15,jli16}, classification \citep{pajares10,zhang17}, optimization \citep{hopfield85,serpen08,cli16}, and to implement associative memories \citep{hopfield82,hassoun97}. Likewise, using the background theory presented in this paper, we believe that HHNNs can be applied in control, computer vision and image processing, to solve optimization problems as well as to implement associative memories designed for the storage and recall of multidimensional data. Indeed, complex-valued and quaternion-valued Hopfield neural networks have been effectively applied to implement associative memories, for example in \citep{jankowski96,isokawa13,isokawa18,tanaka09}. In particular, we \obs{intend} to develop efficient learning rules for HHNNs in the future.}

\section*{Appendix -- Proof of Theorem \ref{thm:HHS}}

First of all, {since $\mathcal{S}$ is compact, the Cartesian product $\mathcal{S}^N$ is also compact.} Furthermore, the function $E$ given by \eqref{eq:Energy} (or, equivalently by \eqref{eq:EnergyB}) is real-valued, by definition. Also, it is continuous because $\mathcal{B}$ is a symmetric bilinear form on a finite dimensional vector space. {Now, a well-known result from calculus for continuous function ensures that the continuous image of a compact set is compact (see Corollary 2.5-7 of \cite{kreyszig89} or Theorem 4.25 of \cite{apostol64}).
Therefore, $E:\mathbb{S}^N \to \mathbb{R}$ attains its maximum and minimum values at some points of $\mathcal{S}^N$. In other words, $E$ is bounded.} Let us now show that $E$ is strictly decreasing along any non-stationary trajectory. 

Aiming to simplify notation, let $\overrightarrowtx \equiv \overrightarrowtx(t)$ and $\overrightarrowtx' \equiv \overrightarrowtx(t+\Delta t)$ for some $t \geq 0$. Since we are considering an asynchronous update mode, let us suppose that only the $\mu$th neuron changed its state at iteration $t$. In other words, we assume that $x_j'=x_j$ for all $j \neq \mu$ and $x_\mu' \neq x_\mu$. In this case, the function $E$ evaluated at $\overrightarrowtx$ and $\overrightarrowtx'$ satisfy:
 \begin{align*}
  E(\overrightarrowtx) &=  
  -\frac{1}{2} \Bigg[ \sum_{i \neq \mu} \sum_{j \neq \mu} \mathcal{B}(x_i,w_{ij} x_j) + \sum_{j \neq \mu} \mathcal{B}(x_\mu, w_{\mu j} x_j) \\ 
   & \qquad  + \sum_{i \neq \mu} \mathcal{B}(x_i,w_{i\mu}x_\mu) + \mathcal{B}(x_\mu,w_{\mu\mu} x_\mu) \Bigg],
 \end{align*}
and, since $x_j' = x_j$ for all $j \neq \mu$, we have
 \begin{align*}
E(\overrightarrowtx') &=  
  -\frac{1}{2} \Bigg[ \sum_{i \neq \mu} \sum_{j \neq \mu} \mathcal{B}(x_i,w_{ij} x_j) + \sum_{j \neq \mu} \mathcal{B}(x_\mu', w_{\mu j} x_j)  \\ 
   & \qquad  + \sum_{i \neq \mu} \mathcal{B}(x_i,w_{i\mu}x_\mu') + \mathcal{B}(x_\mu',w_{\mu\mu} x_\mu') \Bigg].
 \end{align*}
Hence, using the linearity of $\mathcal{B}$, the variation of the energy from time $t$ to $t+\Delta t$ is
\begin{align*}
 \Delta E &=  
  -\frac{1}{2}\Bigg[ \sum_{j \neq \mu} \mathcal{B}(x_\mu'-x_\mu, w_{\mu j} x_j) + \sum_{i \neq \mu} \mathcal{B}(x_i,w_{i\mu}(x_\mu'-x_\mu)) \\ & \qquad + \mathcal{B}(x_\mu',w_{\mu\mu} x_\mu') -  \mathcal{B}(x_\mu,w_{\mu\mu} x_\mu) \Bigg].
\end{align*}
Replacing $i$ by $j$ in the second sum, using the identity \eqref{eq:inner-hermitian}, the symmetry of $\mathcal{B}$, and recalling that $\tau(w_{j\mu}) = w_{\mu j}$, we obtain
\begin{align*}
 \Delta E =  
  -\sum_{j \neq \mu} \mathcal{B}(x_\mu'-x_\mu, w_{\mu j} x_j)  
  -\frac{1}{2} \left(\mathcal{B}(x_\mu',w_{\mu\mu} x_\mu') -  \mathcal{B}(x_\mu,w_{\mu\mu} x_\mu) \right).
\end{align*}
From the linearity and symmetry of $\mathcal{B}$, using \eqref{eq:inner-hermitian} again, and recalling that the activation potential of the $\mu$th neuron at iteration $t$ can be expressed as 
\[v_\mu = \sum_{j=1}^N w_{\mu j}x_j = \sum_{j \neq \mu} w_{\mu j}x_j + w_{\mu\mu} x_\mu,\] 
we have
\begin{align*}
 \Delta E 
  &=-\mathcal{B}\left(x_\mu'-x_\mu, \sum_{j \neq \mu} w_{\mu j} x_j\right)-\frac{1}{2} \left(\mathcal{B}(x_\mu',w_{\mu\mu} x_\mu') 
  - \mathcal{B}(x_\mu,w_{\mu\mu} x_\mu) \right) \\
  &=-\mathcal{B}(x_\mu'-x_\mu, v_\mu - w_{\mu\mu} x_\mu)-\frac{1}{2} \left(\mathcal{B}(x_\mu',w_{\mu\mu} x_\mu') 
  - \mathcal{B}(x_\mu,w_{\mu\mu} x_\mu) \right) \\
  &=-\mathcal{B}(x_\mu'-x_\mu, v_\mu)+\mathcal{B}(x_\mu'-x_\mu,w_{\mu\mu} x_\mu) 
  -\frac{1}{2}\mathcal{B}(x_\mu',w_{\mu\mu} x_\mu') + \frac{1}{2} \mathcal{B}(x_\mu,w_{\mu\mu} x_\mu) \\
  &=-\mathcal{B}(x_\mu'-x_\mu, v_\mu)-\frac{1}{2} \left(\mathcal{B}(x_\mu',w_{\mu\mu} x_\mu') 
  - 2\mathcal{B}(x_\mu',w_{\mu\mu} x_\mu) + \mathcal{B}(x_\mu,w_{\mu\mu} x_\mu) \right) \\
  &=-\mathcal{B}(x_\mu'-x_\mu, v_\mu)-\frac{1}{2} \mathcal{B}(x_\mu'-x_\mu,w_{\mu\mu}(x_\mu'-x_\mu)).
\end{align*}
Now, since we are assuming $f(v_\mu)=x_\mu' \neq x_\mu$, we must have $v_\mu \in \mathcal{D}$. Moreover, since $f$ is a $\mathcal{B}$-projection function and $x_\mu \neq f(v_\mu)$, we obtain 
\begin{align*}
-&\mathcal{B}(x_\mu'-x_\mu, v_\mu) = -\mathcal{B}(x_\mu', v_\mu)+\mathcal{B}(x_\mu, v_\mu) \\&=
-\left(\mathcal{B}(f(v_\mu), v_\mu)-\mathcal{B}(x_\mu, v_\mu)\right)<0.
\end{align*} 
As a consequence, we have
\bb \Delta E < -\frac{1}{2} \mathcal{B}(x_\mu'-x_\mu,w_{\mu\mu}(x_\mu'-x_\mu)).\ee
On the one hand, if $w_{ii}=0$ for all $i=1,\ldots,N$, then \[\Delta E < -(1/2) \mathcal{B}(x_\mu'-x_\mu,0)=0,\] which concludes the proof in the case (a).
On the other hand, if $\mathbb{H}$ is a positive semi-definite real-part associative hypercomplex number system, then $\mathcal{B}(x_\mu'-x_\mu,x_\mu'-x_\mu)\geq 0$. Also, if $w_{\mu \mu}$ is a non-negative real number, then 
\begin{align*}
\Delta E &<-\frac{1}{2}\mathcal{B}(x_\mu'-x_\mu,w_{\mu\mu}(x_\mu'-x_\mu)) 
\leq -\frac{1}{2} w_{\mu\mu} \mathcal{B}(x_\mu'-x_\mu,x_\mu'-x_\mu) \leq 0,
\end{align*}
 which concludes the proof in the case (b).

\section*{Acknowledgment}
This work was supported in part by CNPq under grant no. 310118/2017-4 and FAPESP under grant no. 2019/02278-2.

 \bibliographystyle{model5-names}
 \biboptions{authoryear}


\end{document}